%% file: EvalQReason_Framework_Research.tex
\documentclass[journal]{IEEEtran}

\IEEEoverridecommandlockouts
\usepackage{colortbl}
\usepackage[table]{xcolor}
\usepackage{tabularx}
\usepackage{booktabs}         % professional table lines
\usepackage{tabularx}         % auto column width tables
\usepackage{makecell}         % multi-line cells (for wrapped LLM names)
\usepackage{multirow}         % multirow cells
\usepackage{array}            % extended column alignment

\usepackage[utf8]{inputenc}

\usepackage[caption=false,font=footnotesize]{subfig}

\usepackage{cite}
\usepackage{amsmath,amssymb,amsfonts}
\usepackage{algorithmic}
\usepackage{graphicx}
\usepackage{tikz}        % For custom legend
\usepackage{xcolor}

\definecolor{ieeeblue}{HTML}{0173B2}
\definecolor{ieeesoftblue}{HTML}{6BAED6}
\definecolor{ieeeorange}{HTML}{DE8F05}
\definecolor{ieeecoral}{HTML}{FC8D62}
\definecolor{ieeegreen}{HTML}{66C2A5}

\usepackage{textcomp}
\def\BibTeX{{\rm B\kern-.05em{\sc i\kern-.025em b}\kern-.08em
    T\kern-.1667em\lower.7ex\hbox{E}\kern-.125emX}}
    
\begin{document}

\title{EvalQReason: A Framework for Step-Level Reasoning Evaluation 
in Large Language Models}

%\author{Anonymous Authors}

\author{Shaima Ahmad Freja, Ferhat Ozgur Catak, \IEEEmembership{Senior Member, IEEE}, Betul Yurdem, and Chunming Rong, \IEEEmembership{Senior Member, IEEE}
%\thanks{This paragraph of the first footnote will contain the date on which you submitted your paper for review. It will also contain support information, including sponsor and financial support acknowledgment. For example, ``This work was supported in part by the U.S. Department of Commerce under Grant 123456.'' }
\thanks{Shaima Ahmad Freja, Ferhat Ozgur Catak, and Chunming Rong are with the Department of Electrical Engineering and Computer Science, University of Stavanger, 4021 Stavanger, Norway (e-mail: \{shaima.a.freja, f.ozgur.catak,chunming.rong\}@uis.no).}
\thanks{Betul Yurdem is with the Department of Electrical and Electronics Engineering, Izmir Bakircay University, 35665 Izmir, Turkiye (e-mail: betul.yurdem@bakircay.edu.tr).}}

\maketitle

\begin{abstract}

Large Language Models (LLMs) are increasingly deployed in critical applications requiring reliable reasoning, yet their internal reasoning processes remain difficult to evaluate systematically. Existing methods focus on final-answer correctness, providing limited insight into how reasoning unfolds across intermediate steps.

We present \textit{EvalQReason}, a framework that quantifies LLM reasoning quality through step-level probability distribution analysis without requiring human annotation. The framework introduces two complementary algorithms: Consecutive Step Divergence (CSD), which measures local coherence between adjacent reasoning steps, and Step-to-Final Convergence (SFC), which assesses global alignment with final answers. Each algorithm employs five statistical metrics to capture reasoning dynamics. Experiments across mathematical and medical datasets with open-source 7B-parameter models demonstrate that CSD-based features achieve strong predictive performance for correctness classification, with classical machine learning models reaching F1=0.78 and ROC-AUC=0.82, and sequential neural models substantially improving performance (F1=0.88, ROC-AUC=0.97). CSD consistently outperforms SFC, and sequential architectures outperform classical machine learning approaches.

Critically, reasoning dynamics prove domain-specific: mathematical reasoning exhibits clear divergence-based discrimination patterns between correct and incorrect solutions, while medical reasoning shows minimal discriminative signals, revealing fundamental differences in how LLMs process different reasoning types. EvalQReason enables scalable, process-aware evaluation of reasoning reliability, establishing probability-based divergence analysis as a principled approach for trustworthy AI deployment.

\end{abstract}


\section{Introduction}

Large Language Models (LLMs) are increasingly deployed in critical applications such as medical diagnostics~\cite{singhalLargeLanguageModels2023, savageDiagnosticReasoningPrompts2024}, advanced mathematics~\cite{hendrycks2021measuring, luSurveyDeepLearning2023}, and scientific research. 
Despite their impressive performance, the internal reasoning processes of 
these models remain difficult to interpret and systematically evaluate. While comprehensive surveys have documented various dimensions of LLM 
evaluation~\cite{changSurveyEvaluationLarge2024}, existing reasoning evaluation approaches focus primarily on final-answer accuracy, offering limited insight into how reasoning unfolds across intermediate steps or whether logical coherence is preserved throughout the inference process~\cite{li2025think}. 
This lack of transparency poses significant challenges for trustworthiness and reliability, particularly in high-stakes domains where understanding \textit{how} a model arrives at a conclusion is as important as \textit{whether} the conclusion is correct.
This gap in evaluation fundamentally limits progress toward trustworthy AI. When models produce incorrect solutions, current methods cannot localize where or why the reasoning process failed~\cite{zheng2025processbench}. Moreover, existing frameworks lack a unified mechanism for capturing the dynamic and sequential nature of LLM reasoning using continuous, probability-based signals. 
Consequently, reasoning behavior cannot be systematically analyzed across domains, difficulty levels, and correctness outcomes, hindering principled 
assessment of model reliability in critical applications~\cite{wang2025comprehensive,huangTrustLLMTrustworthinessLarge2024, li2025think}.

We introduce \textbf{EvalQReason}, a three-stage framework for quantitative, probability-based evaluation of LLM reasoning that does not require human annotation or external evaluators. EvalQReason operates by (1) extracting step-level logits from model-generated reasoning chains, 
(2) quantifying reasoning dynamics using two divergence-based algorithms, and (3) performing predictive modeling to assess reasoning reliability. Unlike approaches that depend on labeled reasoning traces~\cite{lightman2023let, zheng2025processbench} 
or semantic evaluation frameworks based on LLM judges~\cite{haoLLMReasonersNew2024b, doWhatDefinesGood2025}, 
EvalQReason derives continuous statistical signals directly from the model’s own probability distributions.

There are two complementary algorithms at the core of the framework. Consecutive Step Divergence (CSD) quantifies local reasoning coherence by measuring distributional shifts between adjacent reasoning steps, while Step-to-Final Convergence (SFC) captures global alignment by comparing intermediate steps to the final-answer distribution. Using five statistical metrics, Kullback–Leibler (KL) Divergence \cite{kullback1951information}, Jensen–Shannon (JS) Divergence \cite{61115}, Hellinger Distance \cite{Hellinger1909}, Cosine Similarity \cite{chowdhury2010introduction}, and Entropy Difference \cite{cover1999elements}, EvalQReason characterizes both local and global properties of reasoning dynamics. Experimental results across mathematical and medical domains reveal that these dynamics are fundamentally domain-specific: divergence-based patterns strongly discriminate correctness in mathematical reasoning, while exhibiting minimal discriminative signal in medical diagnostic tasks.

This work makes two primary contributions:
\begin{enumerate}
    \item We propose EvalQReason, a three-stage framework that quantifies LLM reasoning dynamics from step-level probability distributions without requiring human annotation.
    
    \item We introduce two divergence-based algorithms, Consecutive Step Divergence (CSD) and Step-to-Final Convergence (SFC), that capture local and global reasoning behavior across domains, difficulty levels, and reasoning lengths.
\end{enumerate}

The remainder of this article is organized as follows. 
Section~\ref{sec:related_work} reviews related work on reasoning evaluation and 
interpretability. Section~\ref{sec:methodology} presents the EvalQReason 
framework, including datasets, models, and the CSD/SFC algorithms. 
Section~\ref{sec:results} reports experimental results and analysis. 
Section~\ref{sec:limitations} discusses limitations and future directions, and Section~\ref{sec:conclusion} concludes the paper.

\section{Related Work}
\label{sec:related_work}

The trustworthiness of LLM reasoning is critical as models are deployed in high-stakes domains, motivating evaluation beyond final answer accuracy~\cite{wang2025comprehensive}. Traditional benchmarks that rely solely on outcome-oriented metrics risk masking systemic deficiencies in the reasoning process itself, such as logical fallacies or error propagation~\cite{lee2025evaluating, mondorf2024beyond, li2025think}. 
This has motivated a shift toward process-aware assessment methods.

\subsection{Existing Evaluation Paradigms}

Early approaches trained outcome reward models to verify final answers~\cite{cobbe2021training}, which Lightman et al. extended to process 
supervision by training reward models on 800K+ human annotations to score 
individual steps~\cite{lightman2023let}. Math-Shepherd~\cite{wang2024math} 
and ProcessBench~\cite{zheng2025processbench} further developed process 
reward models for step-level evaluation, but these supervised approaches 
require substantial human labor and struggle with domain transfer beyond 
training data (GSM8K, MATH).

To avoid human annotation, recent frameworks employ LLMs as 
evaluators~\cite{gu2024survey}. ROSCOE~\cite{golovneva2022roscoe} provides 
multi-dimensional evaluation but relies on reference-based comparisons that 
limit accommodation of diverse reasoning paths. ReasonEval~\cite{xia2025evaluating} 
assesses validity and redundancy using trained LLM judges, while 
CaSE~\cite{doWhatDefinesGood2025} evaluates relevance and coherence in a 
reference-free manner. AutoRace~\cite{haoLLMReasonersNew2024b} uses GPT-4 to 
evaluate reasoning chains based on automatically generated rubrics. While 
interpretable, these approaches depend on external LLMs (often proprietary 
models) and provide categorical rather than continuous quantification of 
reasoning dynamics.

Foundational work on Chain-of-Thought prompting~\cite{wei2022chain} and self-consistency methods~\cite{wang2022self} demonstrated how to elicit and improve reasoning chains, though these focus on generation rather than evaluation.

\subsection{Domain-Specific Reasoning Challenges}

LLM reasoning capabilities vary fundamentally across domains. Mathematical 
reasoning requires multi-step logical deduction~\cite{hendrycks2021measuring}, 
while statistical and causal reasoning present distinct 
challenges~\cite{liuAreLLMsCapable2024}. Medical reasoning is particularly problematic: 
Savage et al.~\cite{savageDiagnosticReasoningPrompts2024} demonstrated the importance of 
interpretable diagnostic reasoning, while Wu et al.~\cite{wuWhyChainThought2025} 
showed that chain-of-thought reasoning systematically fails in clinical 
contexts (86.3\% of models show degradation), attributed to the complexity 
and fragmentation of clinical documentation. These domain-specific differences 
motivate evaluation frameworks that systematically characterize how reasoning 
dynamics vary across problem types.

\subsection{Positioning EvalQReason}

While existing approaches provide valuable evaluation frameworks, they face key limitations. Supervised methods~\cite{lightman2023let,wang2024math,zheng2025processbench} 
require extensive annotation and struggle with domain transfer. LLM-judge 
approaches~\cite{xia2025evaluating,doWhatDefinesGood2025,haoLLMReasonersNew2024b} 
depend on external models and provide categorical assessments.

EvalQReason addresses these limitations through fully automated, probability-based 
divergence analysis. By quantifying step-level reasoning dynamics through 
probability distributions—signals that recent work has shown reflect genuine 
model confidence~\cite{kadavath2022language} and enable effective 
adaptation~\cite{hiranandaniLogitsAreAll2025}—our framework requires no human 
supervision or external evaluators. As demonstrated in 
Section~\ref{subsubsec:comparison}, this approach outperforms existing supervised 
and LLM-judge methods while eliminating annotation costs entirely.

\section{Methodology}
\label{sec:methodology}
 
The methodology begins with the use of mathematical and medical datasets combined with domain-specialized models (Table~\ref{tab:datasets}) to establish the experimental foundation. As illustrated in Figure~\ref{fig:framework}, the framework consists of three main stages: 
(i) generating reasoning steps and extracting token-level probabilities, 
(ii) evaluating reasoning coherence and convergence through the Consecutive Step Divergence (CSD) and Step-to-Final Convergence (SFC) algorithms, and 
(iii) performing pattern analysis and predictive modeling to assess reasoning reliability.

\subsection{Datasets and Models}
\label{subsec:datasets_models}

To analyze reasoning reliability across domains, we employ three challenging benchmarks: AIME, Math-500, and MedQA, covering mathematical question answering (QA) and medical multiple-choice questions (MCQ). Mathematical reasoning is a critical testbed for LLM evaluation~\cite{hendrycks2021measuring}, 
as it requires precise multi-step deduction and error-sensitive computation. 
These properties make math datasets particularly suitable for process-level analysis of reasoning dynamics across intermediate steps. Medical reasoning provides a complementary setting, emphasizing domain knowledge and diagnostic inference, enabling cross-domain comparison of reasoning coherence under fundamentally different constraints.

The selected datasets are designed to span diverse reasoning patterns, 
difficulty levels, and domain-specific knowledge requirements. Dataset statistics and difficulty distributions are summarized in Table~\ref{tab:datasets}.
\input{tables/datasets.tex}

\noindent\textbf{1) AIME:}
A collection of 935 high-difficulty mathematical problems from the \textit{American Invitational Mathematics Examination} (1983--2024).
These problems are designed to assess advanced mathematical reasoning through creative, multi-step problem solving.
We use a cleaned version derived from the publicly available Kaggle release~\cite{READMEmdGneubigAime198320242024}, after removing duplicates and incomplete entries.

\textit{Difficulty Reclassification.} To support reasoning-level analysis, the original 15-problem ordering in each exam was aggregated into three difficulty levels: Easy (Problems 1–5), Medium (6–10), and Hard (11–15). This grouping reflects the inherent progression of problem difficulty and enables controlled comparison across reasoning depths. To mitigate class imbalance, we apply stratified sub-sampling, yielding a balanced subset of 240 problems (80 per difficulty level). Details are provided in Section~\ref{subsec:stage1}.

\noindent\textbf{2) Math-500:} A curated subset of 500 competition-level problems from the MATH benchmark~\cite{hendrycks2021measuring}. 
The original MATH dataset was introduced to evaluate multi-step mathematical reasoning in large language models and contains over 12,000 problems spanning five difficulty levels. 
This subset preserves the same structure (Levels~1--5) and exhibits a natural skew toward greater difficulty, making it well-suited for fine-grained analysis of reasoning stability under increasing problem complexity.

\noindent\textbf{3) MedQA:}
A large-scale medical multiple-choice question answering benchmark constructed from professional medical board examinations~\cite{jinWhatDiseaseDoes2021}.
It evaluates models’ ability to perform clinical reasoning and apply domain knowledge to diagnostic decision-making.
We use the English subset to ensure consistency with the mathematical datasets, while providing a contrasting domain for analyzing differences between structured problem solving and knowledge-intensive inference.

\textit{Preprocessing and Difficulty Grouping.} 
Difficulty levels were derived from the \texttt{meta\_info} field, which 
categorizes questions into Type~1 (single knowledge point) and Type~2 
(multi-step clinical scenario). These categories were mapped to two 
difficulty levels: Level~1 (Basic) and Level~2 (Advanced), enabling analysis of reasoning behavior across increasing clinical complexity.
\subsection{Models}
We employ three open-source 7B-parameter LLMs to generate step-by-step reasoning chains across mathematical and medical domains. The use of open-source models is methodologically essential for EvalQReason, as the framework requires direct access to token-level logits at inference time to extract step-level probability distributions for divergence analysis. Such access is unavailable in closed-source or API-based models (e.g., GPT-4, Gemini), which expose only final text outputs and therefore cannot support process-level reasoning analysis based on internal 
distributional dynamics.

\begin{itemize}
    \item \textbf{Mathstral~7B}~\cite{MathStralMistralAI}: A Mistral~7B variant fine-tuned on synthetic mathematical data, exhibiting strong multi-step logical reasoning suitable for AIME and Math-500.
    \item \textbf{Qwen2.5~7B-Instruct}~\cite{yangQwen2TechnicalReport2024}: The primary model used across all datasets, selected for its strong instruction-following capability, enhanced mathematical reasoning, and extended context length (up to 128K tokens).
    \item \textbf{Qwen-Medicine~7B}~\cite{WangCaQwen257BMedicineHugging}: A medical-domain extension of Qwen2.5~7B fine-tuned on 340K medical dialogues, optimized for clinical terminology and diagnostic reasoning.
\end{itemize}

These models enable controlled comparison of reasoning dynamics across 
mathematical and medical problem-solving domains.

\subsection{EvalQReason Framework Architecture}
\label{subsec:framework_overview}

\begin{figure*}[t]
  \centering
  \includegraphics[width=0.95\textwidth]{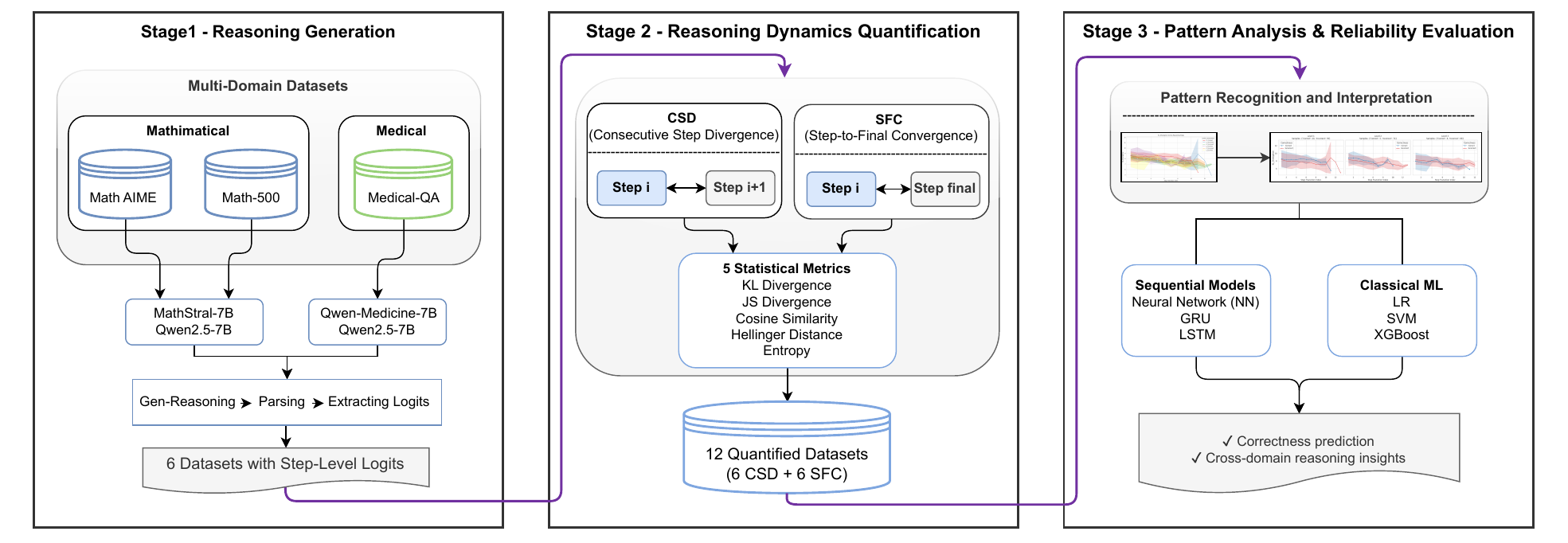}
  \caption{EvalQReason framework architecture.}
  \label{fig:framework}
\end{figure*}

EvalQReason is a three-stage framework for analyzing and quantifying reasoning behavior in LLMs. As illustrated in Figure~\ref{fig:framework}, the framework first generates step-by-step reasoning chains and extracts token-level logits (Stage~1). These logits are transformed into step-level probability distributions, which are then quantified using statistical divergence metrics to capture local coherence and global convergence across reasoning steps (Stage~2). Finally, EvalQReason performs pattern analysis and predictive 
modeling to assess reasoning reliability and characterize domain-specific dynamics (Stage~3). This modular design enables systematic analysis of how reasoning evolves across steps and how evaluation behavior varies across domains and difficulty levels.

\subsubsection{Stage 1: Reasoning Generation and Logit Extraction}
\label{subsec:stage1}

Stage 1 establishes the foundation for quantitative reasoning analysis by generating structured reasoning chains and extracting step-level logit distributions from LLMs across diverse domains and difficulty levels. The output is a curated dataset containing reasoning chains with their corresponding raw logit matrices, forming the input for divergence-based analysis in Stage~2.

We employ domain-specialized model pairs: (i)~MathStral-7B and Qwen2.5-7B for mathematical reasoning (AIME, Math-500), and (ii)~Qwen2.5-7B and Qwen-Medicine-7B for medical reasoning (MedQA). Domain-specific prompts (Supplementary Material, Fig. S1) encourage structured step-by-step reasoning with explicit formatting requirements. For mathematical problems, models generate clear logical steps culminating in a boxed final answer. For medical problems, models work through clinical cases by analyzing symptoms, eliminating incorrect options with justification, and selecting the correct answer. Reasoning chains are generated using greedy decoding (temperature=0) to ensure deterministic outputs, with logit collection enabled at each generation step.

The pipeline consists of four sequential steps:

\textbf{1) Reasoning Chain Generation.}
For each problem, we employ greedy decoding (\texttt{do\_sample=False}) with adaptive token limits of 512, 1024, and 2048 tokens and a repetition penalty of 1.2. Generation proceeds with logit extraction enabled through \texttt{return\_dict\_in\_generate=True} and \texttt{output\_scores=True}. The generated text is validated to confirm the presence of a final answer in the expected format (e.g., \texttt{\textbackslash boxed\{\}} for mathematics). If validation fails, generation retries with the next higher token limit.

\textbf{2) Parsing and Step Alignment.}
The generated reasoning text is parsed into individual reasoning steps based on explicit step markers (e.g., ``Step~1:'', ``Step~2:''). Each step's tokens are aligned with their corresponding model-generated logits, establishing a direct mapping between textual reasoning and the model's internal logit outputs.

\textbf{3) Logit Extraction and Storage.}
For each reasoning step $i$ containing $T_i$ tokens, we extract the raw logit matrix $\ell_i \in \mathbb{R}^{T_i \times |V|}$, where each row $\ell_{i,t} \in \mathbb{R}^{|V|}$ contains the logit vector for token $t$ at step $i$, and $|V|$ is the vocabulary size. These raw logit matrices are stored without further processing, preserving the complete distributional information for subsequent analysis.

\textbf{4) Dataset Curation and Storage.}
All reasoning steps and their corresponding logit matrices are stored in a pickle format, preserving problem metadata (difficulty level, domain, correctness label) alongside reasoning dynamics. For AIME, we applied stratified sub-sampling after generating chains for all 935 problems using MathStral-7B. To address severe class imbalance (25.8\% correct), we selected a balanced subset of 240 problems (80 per difficulty level), enabling fair comparison across difficulty strata. For Math-500 and MedQA, natural correctness distributions were preserved without sub-sampling.

This stage produces curated datasets containing parsed reasoning chains with step-level logits, forming the foundation for the quantitative divergence analysis conducted in Stage~2. Dataset statistics and model accuracy are reported in Table~\ref{tab:stage1_dataset}.

\subsubsection{Stage 2: Reasoning Dynamics Quantification}
\label{subsec:stage2}

Stage~2 quantifies reasoning dynamics from the logit representations extracted in Stage~1 using divergence-based analysis. We introduce two complementary 
algorithms: Consecutive Step Divergence (CSD), which measures local coherence between adjacent reasoning steps, and Step-to-Final Convergence (SFC), which assesses global alignment of intermediate steps toward the final answer.

This formulation is inspired by interpretability methods that analyze prediction evolution within LLMs. The logit lens~\cite{nostalgebraistInterpretingGPTLogit2020} 
and tuned lens~\cite{belroseElicitingLatentPredictions2025} project intermediate transformer layers into vocabulary space to study convergence across layers. 
EvalQReason adapts this principle from an architectural setting (layer-wise evolution) to a behavioral setting, quantifying how probability distributions evolve across multi-step reasoning chains to capture both local coherence and global convergence.

\textbf{(i) Step-Level Probability Distribution Construction}

Given raw logit matrices from Stage~1, we construct step-level probability distributions through token-level normalization and averaging. For each reasoning step $i$ with logit matrix $\ell_i \in \mathbb{R}^{T_i \times |V|}$, we apply softmax to each token's logit vector:
\begin{equation}
p_{i,t} = \text{softmax}(\ell_{i,t}) \in \Delta^{|V|-1}, \quad t = 1,\ldots,T_i
\end{equation}
The step-level distribution is the arithmetic mean across all tokens:
\begin{equation}
\bar{p}_i = \frac{1}{T_i}\sum_{t=1}^{T_i} p_{i,t} \in \Delta^{|V|-1}
\end{equation}
This yields a mixture distribution representing typical next-token preferences within the step, ensuring length invariance and preserving full vocabulary information. For numerical stability, we apply additive smoothing $\tilde{p}_i = \bar{p}_i + \varepsilon$ where $\varepsilon = 10^{-7}$. All divergences are computed over the full vocabulary using natural logarithms (units: nats).

\textbf{(ii) Divergence-Based Algorithms}

Using the step-level distributions $\{\tilde{p}_1, \tilde{p}_2, \ldots, \tilde{p}_n\}$, we quantify reasoning dynamics through two complementary algorithms.

\textit{Consecutive Step Divergence (CSD).}
CSD measures local reasoning coherence by quantifying distributional shifts 
between adjacent steps. For a reasoning chain with $n$ steps, we compute the divergence between each consecutive pair:
\begin{equation}
\text{CSD}_i = D(\tilde{p}_i \| \tilde{p}_{i+1}), \quad i = 1, 2, \ldots, n-1
\end{equation}
where $D(\cdot, \cdot)$ represents a divergence metric. Low CSD values 
indicate smooth reasoning progression; high values suggest reasoning drift 
or contradictions.

\textit{Step-to-Final Convergence (SFC).}
SFC evaluates global alignment by measuring how each intermediate step 
relates to the final-answer distribution:
\begin{equation}
\text{SFC}_i = D(\tilde{p}_i \| \tilde{p}_{\text{final}}), \quad i = 1, 2, \ldots, n-1
\end{equation}
where $\tilde{p}_{\text{final}}$ is the distribution at the final reasoning 
step. Decreasing SFC values indicate progressive convergence; fluctuating values suggest tangential reasoning.

\textbf{(iii) Statistical Metrics}

We employ five complementary statistical metrics to quantify distributional relationships in both CSD and SFC: Kullback-Leibler (KL) Divergence, Jensen-Shannon (JS) Divergence, Hellinger Distance, Cosine Similarity, and Entropy Difference. These metrics capture distinct aspects of reasoning dynamics—divergence measures (KL, JS, Hellinger) quantify distributional shifts, Cosine Similarity measures directional consistency, and Entropy Difference tracks uncertainty changes. All metrics are computed over the full vocabulary using natural logarithms (units: nats). Complete mathematical definitions are provided in Supplementary Material Section S2.

For CSD, we compute the signed entropy change between consecutive steps:
\begin{equation}
\Delta H_i = H(\mathbf{P}_{i+1}) - H(\mathbf{P}_i)
\label{eq:entropy_csd}
\end{equation}
Positive values indicate increasing uncertainty (model becoming less confident), 
while negative values indicate decreasing uncertainty (model becoming more 
confident).

For SFC, we compute the absolute entropy deviation from the final step:
\begin{equation}
\Delta H_i = \left| H(\mathbf{P}_i) - H(\mathbf{P}_{\text{final}}) \right|
\label{eq:entropy_sfc}
\end{equation}
This measures how each step's uncertainty differs from the final answer's 
uncertainty, with lower values indicating convergence in confidence levels.

The resulting outputs form two structured datasets per dataset-model pair: 
one for CSD and one for SFC. Each CSD dataset contains all five statistical metrics computed between consecutive steps, while each SFC dataset contains the same five metrics computed relative to the final step. Across six dataset-model pairs (AIME-MathStral, AIME-Qwen2.5, Math500-MathStral, Math500-Qwen2.5, MedQA-Qwen2.5, MedQA-QwenMedicine), this produces 12 quantified datasets (6 CSD + 6 SFC), each enriched with metadata including model name, question identifier, difficulty level, and correctness label. 
These datasets serve as the analytical foundation for Stage~3, enabling detailed investigation of reasoning trajectories, domain-specific behavior, and predictive modeling of reliability.

\subsubsection{Stage~3: Pattern Analysis and Reliability Evaluation}
\label{subsec:stage3}

We perform two complementary analyses on the quantified outputs from 
Stage~2: (i) pattern analysis to identify systematic differences in reasoning 
dynamics, and (ii) predictive modeling to evaluate whether these patterns can 
reliably predict answer correctness.

\paragraph{Pattern Recognition and Interpretation.}

To examine how reasoning behavior varies across domains, difficulty levels, and correctness outcomes, we analyze statistical metrics (KL Divergence, JS Divergence, Cosine Similarity, Hellinger Distance, and Entropy Difference) 
stratified by correctness and difficulty. This analysis characterizes how reasoning coherence evolves, fluctuates, or degrades across reasoning steps, providing insight into model reliability and stability.

We study patterns along five dimensions: (i) metric comparison to identify the most discriminative measures, (ii) difficulty-level analysis to assess the impact of problem complexity, (iii) step-length analysis to evaluate the effect of reasoning chain length, (iv) algorithmic comparison (CSD vs.\ SFC) to contrast 
local and global dynamics, and (v) cross-domain analysis to reveal domain-specific reasoning characteristics.

\paragraph{Predictive Modeling of Reasoning Reliability}

To assess whether quantified reasoning dynamics can predict answer correctness, we employ two complementary modeling approaches.

\textbf{(i) Classical Machine Learning Models.}
We train three classical models—Logistic Regression (LR), Support Vector Machine 
(SVM), and XGBoost (XGB)—on engineered temporal features. For each 
dataset–LLM–algorithm combination, we extract approximately 20 features per reasoning chain that capture three aspects of reasoning dynamics: (i) aggregate 
statistics (e.g., mean and final values), (ii) temporal progression (slopes 
reflecting stability trends), and (iii) volatility indicators (maximum changes between consecutive steps).

For divergence-based metrics (KL, JS, and Hellinger), we compute the mean, 
temporal slope, maximum jump, and final value. For Cosine Similarity, we extract 
the mean, slope, and final value. For Entropy Difference, we derive six features 
capturing uncertainty dynamics, including cumulative entropy change and overall 
entropy trends. Together, these features map variable-length reasoning sequences 
to fixed-dimensional representations while preserving both statistical and 
temporal characteristics.

\textbf{(ii) Sequential Modeling of Reasoning Dynamics.}
To capture temporal dependencies in reasoning sequences, we employ Neural 
Network (NN), Gated Recurrent Unit (GRU)~\cite{choLearningPhraseRepresentations2014a}, 
and Long Short-Term Memory (LSTM)~\cite{hochreiterLongShortTermMemory1997a} models. Unlike classical models based on engineered features, these sequential models operate directly on step-wise metric sequences. Each reasoning chain is represented as a temporal sequence $\mathbf{S} = [s_1, s_2, \ldots, s_n]$, where $s_i \in \mathbb{R}^5$ denotes the five-dimensional metric vector at step $i$. Variable-length sequences are handled via dynamic padding within each batch.

All sequential models share a common base architecture with hidden dimensions of 64 and 128, ReLU activation, batch size 32, and training for up to 100 epochs with early stopping (patience = 50). Preserving temporal order enables these models to capture reasoning dynamics such as mid-step instability, progressive degradation, and late-stage convergence that aggregated features cannot capture.

Models are evaluated using stratified train–test splits based on the correctness label. Hyperparameters for classical models are optimized via grid search, randomized search, and Bayesian optimization, while sequential models explore learning rate, L2 regularization, and dropout. The best configuration for each dataset–model–algorithm pair is selected using F1-Score. Performance is reported using F1-Score (primary), ROC-AUC, Accuracy, and Balanced Accuracy for both CSD and SFC. Complete hyperparameters are in Supplementary Material Tables S1-S4.

\section{Results and Discussion}
\label{sec:results}

This section presents experimental findings from applying the EvalQReason framework to evaluate reasoning dynamics across mathematical and medical domains. We first analyze systematic patterns in reasoning behavior (Section~\ref{subsec:pattern_recognition}), revealing how local coherence and global convergence distinguish correct from incorrect reasoning. We then evaluate predictive performance (Section~\ref{subsec:predictive_modeling}), demonstrating that divergence-based features enable reliable correctness prediction.
Table~\ref{tab:stage1_dataset} summarizes the six curated reasoning datasets generated in Stage~1, showing overall correctness distribution and baseline accuracies. Mathematical reasoning proves more challenging than medical reasoning, with AIME presenting the greatest difficulty (14.1--25.8\% accuracy) compared to Math-500 (43.4--48.3\%) and MedQA (42.6--47.6\%), reflecting its competition-level problem complexity.
\input{tables/stage1_dataset.tex}

\subsection{Pattern Recognition and Interpretation}
\label{subsec:pattern_recognition}

We analyze systematic patterns within the quantified CSD and SFC datasets to examine how coherence and convergence behaviors vary across domains, difficulty levels, and correctness outcomes.

\subsubsection{CSD Analysis Across Metrics}
\label{subsubsec:csd_metrics}

\begin{figure*}[!t]
\centering
% ---------- Shared Y-axis label ----------
\makebox[2pt][l]{%
  \raisebox{0.5\height}{%
    \rotatebox{90}{\small Statistical Metric}%
  }%
}
% Row 1
\subfloat[KL Divergence]{
    \includegraphics[width=0.3\textwidth]{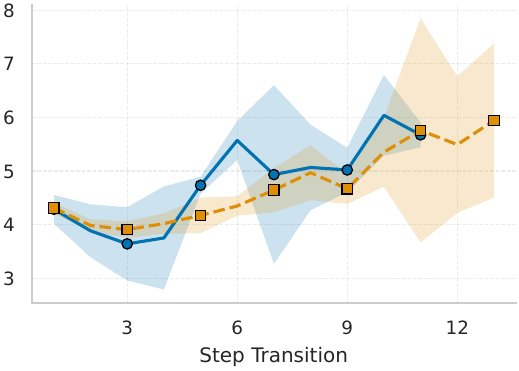}
    \label{fig:kl}
}
\hfill
\subfloat[JS Divergence]{
    \includegraphics[width=0.28\textwidth]{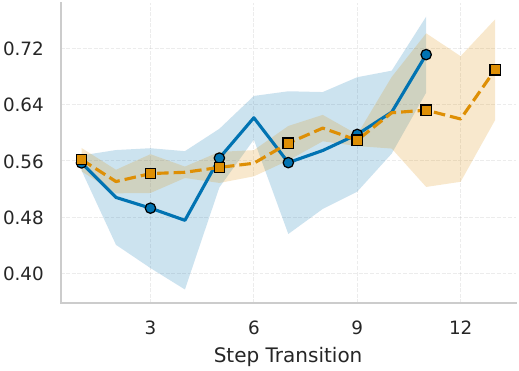}
    \label{fig:js}
}
\hfill
\subfloat[Hellinger Distance]{
    \includegraphics[width=0.28\textwidth]{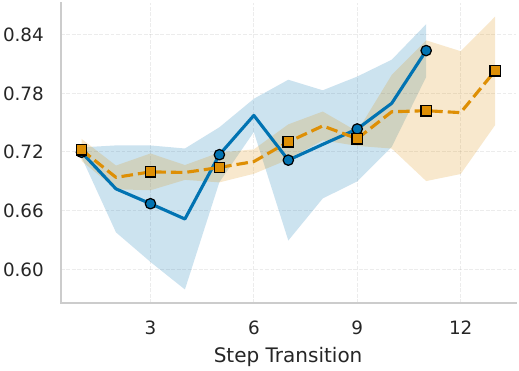}
    \label{fig:hellinger}
}

\vspace{0.3cm}  % Space between rows

% ===== ROW 2: Two subfigures (CENTERED) 
\centering
\makebox[2pt][l]{%
  \raisebox{0.5\height}{%
    \rotatebox{90}{\small Statistical Metric}%
  }%
}
\subfloat[Cosine Similarity]{
    \includegraphics[width=0.28\textwidth]{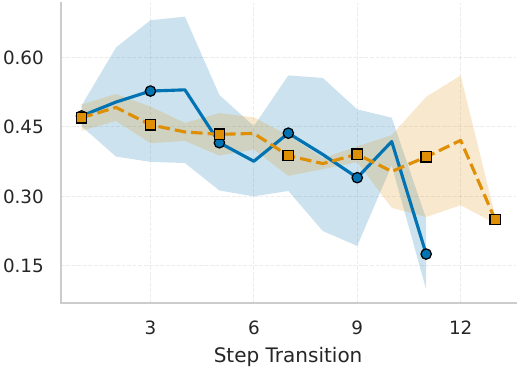}
    \label{fig:cosine}
}
\hspace{0.3cm}
\subfloat[Entropy-Diff]{
    \includegraphics[width=0.28\textwidth]{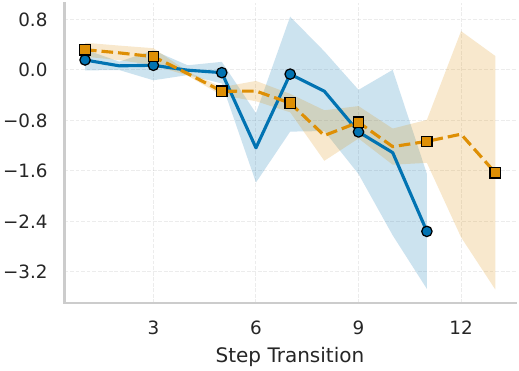}
    \label{fig:entropy}
}

% Shared Legend
\vspace{0.1cm}
\begin{center}
{\footnotesize
\textcolor{ieeeblue}{\rule{1.0em}{2pt}} Correct \quad
\textcolor{ieeeorange}{\rule{1.0em}{2pt}} Incorrect
}
\end{center}
\caption{Comparison of five statistical metrics for CSD computation on AIME (Qwen2.5-7B). 
(a) KL Divergence, (b) JS Divergence, (c) Hellinger Distance, (d) Cosine Similarity, 
and (e) Entropy Difference. Shaded regions: ±1 SD.}
\label{fig:csd_metrics}
\end{figure*}

To identify the most discriminative measure of local reasoning coherence, 
we compare five statistical metrics within the CSD framework. Figure~\ref{fig:csd_metrics} 
compares these metrics for the AIME dataset using Qwen2.5-7B. 

Among the five metrics, KL Divergence demonstrates the 
clearest and most consistent separation between correct and incorrect 
reasoning trajectories, maintaining stability across reasoning steps and 
exhibiting the narrowest confidence intervals. JS Divergence and Hellinger Distance reveal similar but weaker discrimination, particularly 
in mid-step regions. Cosine Similarity exhibits the expected inverse pattern: 
correct reasoning maintains higher similarity values (indicating consistent 
distributional alignment), while incorrect reasoning displays lower similarity 
with greater volatility. Entropy Difference displays higher variability and 
lacks consistent directionality across correctness classes.

This comparison validates KL Divergence as the primary metric for subsequent 
analyses based on its superior sensitivity and interpretability.

\subsubsection{CSD Analysis Across Difficulty Levels and Correctness}
\label{subsubsec:csd_difficulty}

\begin{figure*}[t]
\centering
\begin{tabular}{ccc}
% Row 1: KL Divergence
\includegraphics[width=0.30\textwidth]{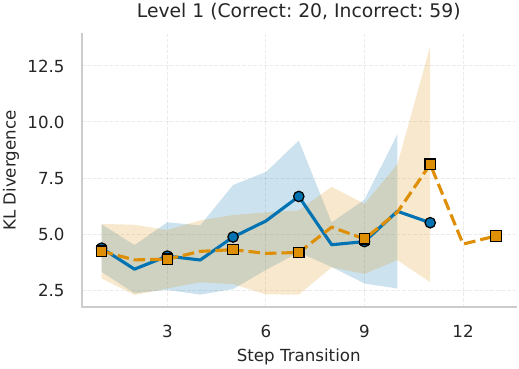} & 
\includegraphics[width=0.30\textwidth]{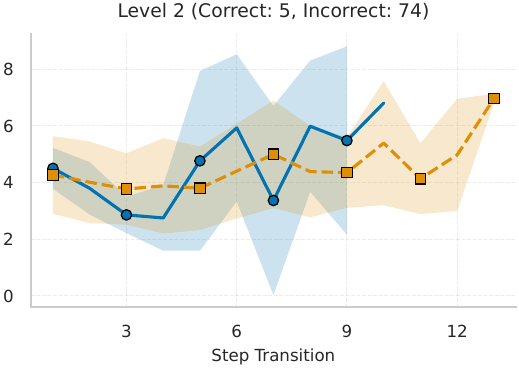} &
\includegraphics[width=0.30\textwidth]{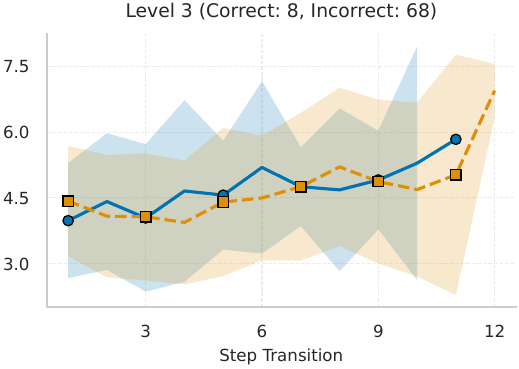} \\

\end{tabular}
% Shared Legend
\vspace{0.1cm}
\begin{center}
{\footnotesize
\textcolor{ieeeblue}{\rule{1.0em}{2pt}} Correct \quad
\textcolor{ieeeorange}{\rule{1.0em}{2pt}} Incorrect
}
\end{center}
\caption{CSD analysis across difficulty levels for the \textit{AIME} dataset using \textit{Qwen2.5-7B} with \textit{KL Divergence}. Panels correspond to Level~1 (Easy), Level~2 (Medium), and Level~3 (Hard).}
\label{fig:aime_csd_levels}
\end{figure*}

We analyze how consecutive-step coherence varies across problem difficulty 
and correctness outcomes using the AIME dataset with Qwen2.5-7B. As shown in Figure~\ref{fig:aime_csd_levels}, correct reasoning consistently maintains lower and more stable divergence values compared to incorrect reasoning across all three difficulty levels. At Level 1 (Easy), correct reasoning exhibits a KL divergence of approximately 4-6 units while incorrect reasoning ranges from 6-10 units. As difficulty increases to Levels 2 and 3, both correct and incorrect trajectories show elevated divergence values and substantially wider confidence intervals, indicating that reasoning stability degrades with task complexity. The persistent separation between correct and incorrect reasoning across all difficulty levels demonstrates that CSD effectively captures reasoning quality differences even as problem complexity increases.

\subsubsection{Reasoning Step-Length Analysis}
\label{subsubsec:step_length_analysis}

To examine the effect of reasoning-chain length on performance and divergence patterns, we group problems into Short, Medium, and Long categories using dataset-adaptive thresholds that reflect each dataset's natural step-count distribution (Table~\ref{tab:step_group_results}). This normalization is necessary because mathematical problems typically require more reasoning steps than medical problems.

Table~\ref{tab:step_group_results} reveals strong domain-dependent behavior. In mathematical reasoning, accuracy consistently declines as chain length increases. For Math-500 with Qwen2.5-7B, accuracy drops from 59.81\% (Short) to 31.11\% (Long), a 48\% relative decrease, while AIME with MathStral-7B shows an even steeper decline from 34.69\% to 13.64\% (61\%). This monotonic degradation is consistent across both mathematical datasets and models. In contrast, medical reasoning on MedQA exhibits qualitatively different behavior. Qwen-7B improves accuracy on longer chains ($51.13\% \rightarrow 58.21\%$), while Qwen-Med-7B remains relatively stable ($39.17\% \rightarrow 44.38\%$), indicating robustness to extended reasoning.

\begin{figure*}[t]
\centering
\begin{tabular}{ccc}
% Row 1: KL Divergence
\includegraphics[width=0.30\textwidth]{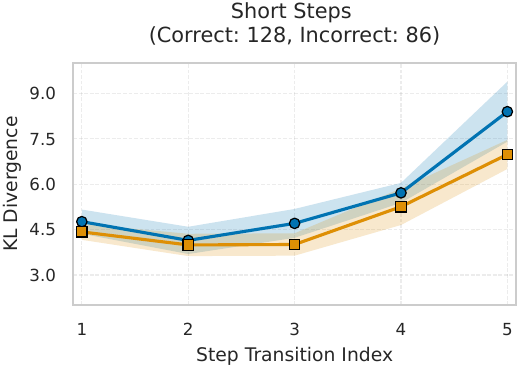} &
\includegraphics[width=0.30\textwidth]{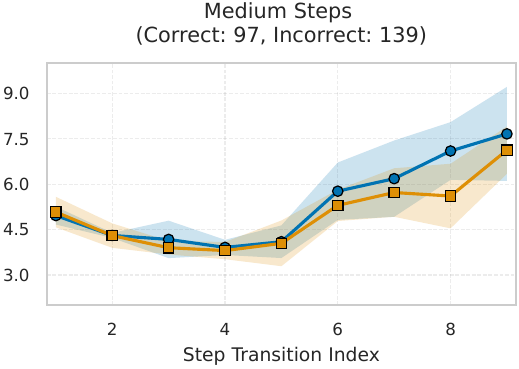} &
\includegraphics[width=0.30\textwidth]{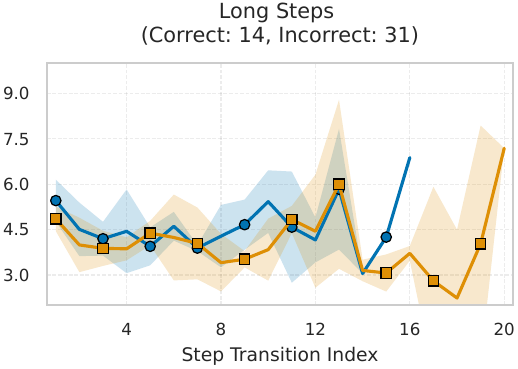} \\
\end{tabular}
% Shared Legend
\vspace{0.1cm}
\begin{center}
{\footnotesize
\textcolor{ieeeblue}{\rule{1.0em}{2pt}} Correct \quad
\textcolor{ieeeorange}{\rule{1.0em}{2pt}} Incorrect
}
\end{center}
\caption{Consecutive Step Divergence (CSD) patterns across step-length 
categories for the Math-500 dataset using Qwen2.5-7B with KL Divergence. 
Step-length categories are defined by total reasoning steps: Short 
($\leq 6$ steps), Medium ($7$--$10$ steps), Long ($\geq 11$ steps).}
\label{fig:step_length_csd}
\end{figure*}

Figure~\ref{fig:step_length_csd} explains the mathematical degradation through Consecutive Step Divergence (CSD) dynamics. Short and medium chains exhibit lower and more stable divergence for correct reasoning, whereas long chains show highly volatile trajectories with widened confidence bands, indicating loss of local coherence. This instability directly accounts for the observed accuracy decline as reasoning length increases.

Overall, these results demonstrate that extended reasoning amplifies instability in mathematical problem-solving, while medical diagnostic reasoning follows distinct dynamics in which longer chains do not necessarily harm—and may even improve—performance.

\input{tables/step_group_results}

\subsubsection{SFC vs. CSD Across Difficulty Levels}
\label{subsubsec:sfc_csd_comparison}

\begin{figure*}[t]
\centering
 % (a) CSD plot
\subfloat[CSD for Math-500 Mathstral (KL Divergence) across difficulty levels.]{%
\includegraphics[width=0.19\textwidth]{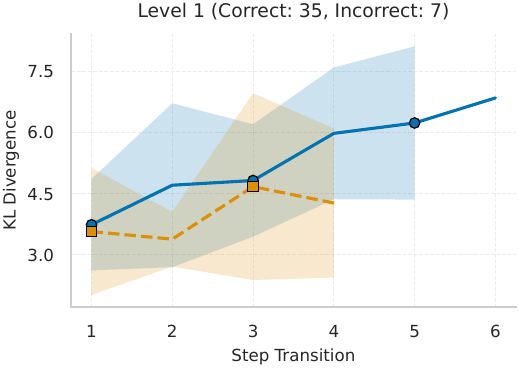}
\hfill
\includegraphics[width=0.19\textwidth]{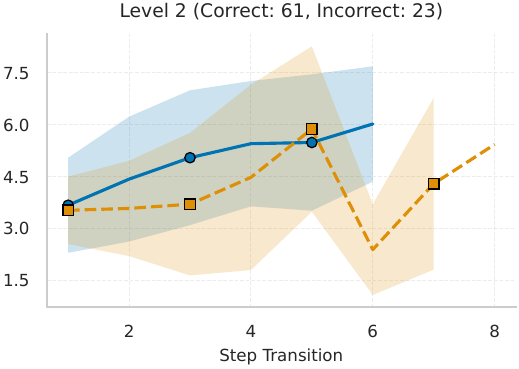}
\hfill
\includegraphics[width=0.19\textwidth]{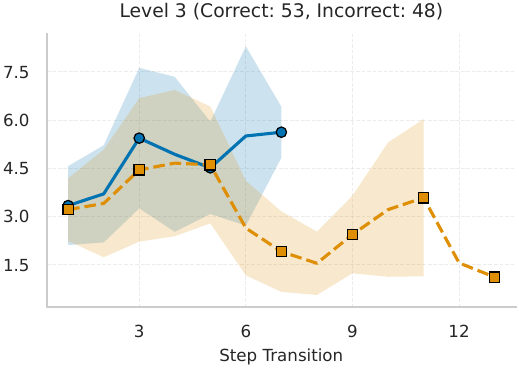}
\hfill
\includegraphics[width=0.19\textwidth]{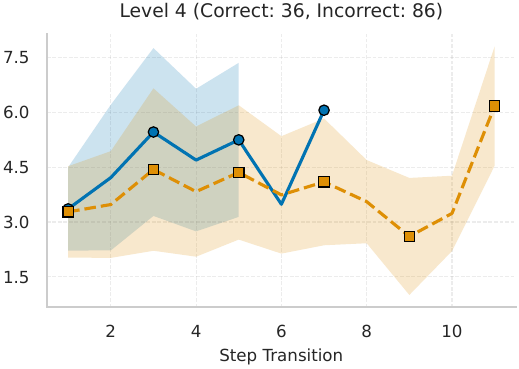}
\hfill
\includegraphics[width=0.19\textwidth]{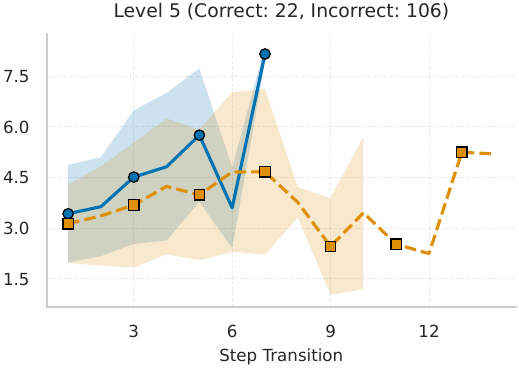}
 \label{fig:math500_csd}%
  }\par\medskip
% Row 2
 \subfloat[SFC for Math-500 Mathstral (KL Divergence) across difficulty levels.]{%
\includegraphics[width=0.19\textwidth]{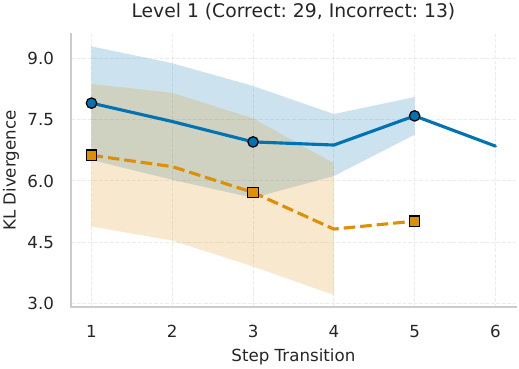}
\includegraphics[width=0.19\textwidth]{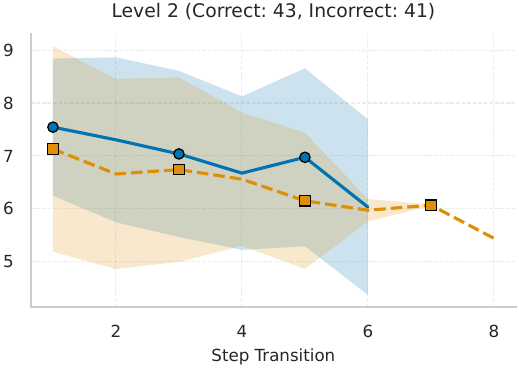}
\includegraphics[width=0.19\textwidth]{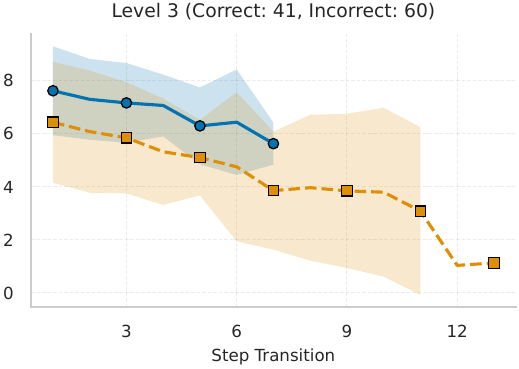}
\includegraphics[width=0.19\textwidth]{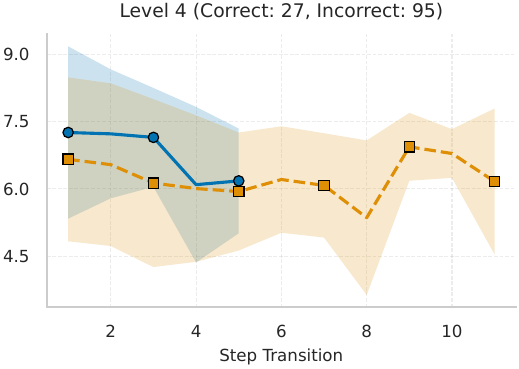}
\includegraphics[width=0.19\textwidth]{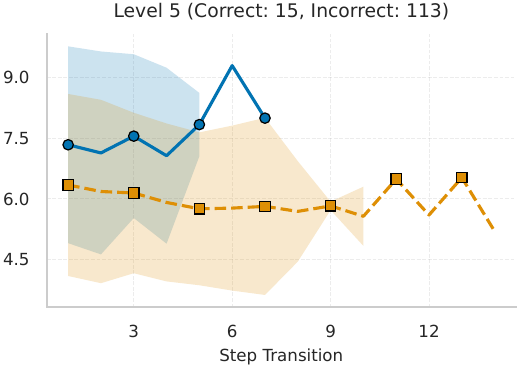} \label{fig:math500_sfc}%
}
% Shared Legend
\vspace{0.1cm}
\begin{center}
{\footnotesize
\textcolor{ieeeblue}{\rule{1.0em}{2pt}} Correct \quad
\textcolor{ieeeorange}{\rule{1.0em}{2pt}} Incorrect
}
\end{center}

 \caption{Comparison of \textit{Consecutive Step Divergence (CSD)} and 
  \textit{Step-to-Final Convergence (SFC)} using \textit{KL Divergence} 
  across five difficulty levels in the \textit{Math-500} dataset with 
  \textit{Mathstral-7B}. (a) CSD measures local coherence with varying correctness separation. (b) SFC measures global convergence toward the final answer with minimal correctness separation.}  
  \label{fig:csd_sfc_comparison}
\end{figure*}

To compare how local coherence (CSD) and global convergence (SFC) capture 
reasoning quality, Figure~\ref{fig:csd_sfc_comparison} presents both algorithms on Math-500 with MathStral-7B across five difficulty levels.
CSD (Fig.~\ref{fig:csd_sfc_comparison}a) shows varying separation between 
correct (blue) and incorrect (orange) reasoning, with incorrect trajectories 
displaying greater volatility, particularly at higher difficulty levels. In 
contrast, SFC (Fig.~\ref{fig:csd_sfc_comparison}b) displays consistently monotonic decrease for both correctness groups, reflecting gradual convergence toward the final answer regardless of correctness. Both correct and incorrect 
reasoning follow nearly parallel trajectories with overlapping confidence 
intervals.
This algorithmic comparison reveals that CSD exposes local structural 
inconsistencies that distinguish correct from incorrect reasoning, while SFC captures global convergence strength that occurs similarly for both. 
This fundamental difference has direct implications for predictive modeling (Section~\ref{subsec:predictive_modeling}), where the ability to detect local coherence breakdowns becomes critical for correctness discrimination.

\subsubsection{Cross-Domain Comparison}
\label{subsubsec:cross_domain}

\begin{figure*}[!t]
  \centering

  % Row 1: AIME (3 panels)
  \subfloat[AIME -- MathStral-7B (KL Divergence) across difficulty levels.]{%
    \makebox[0.99\textwidth][c]{%
      \includegraphics[width=0.27\textwidth]{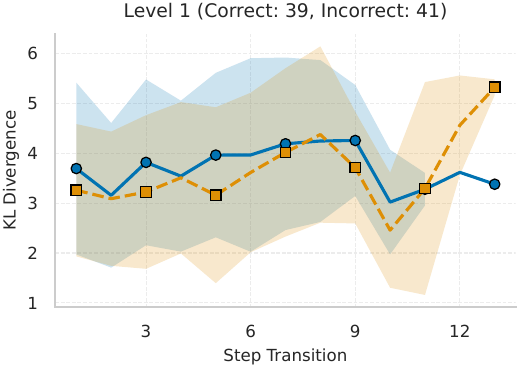}
      \includegraphics[width=0.27\textwidth]{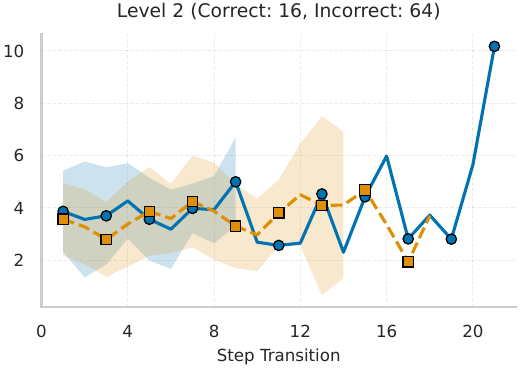}
      \includegraphics[width=0.27\textwidth]{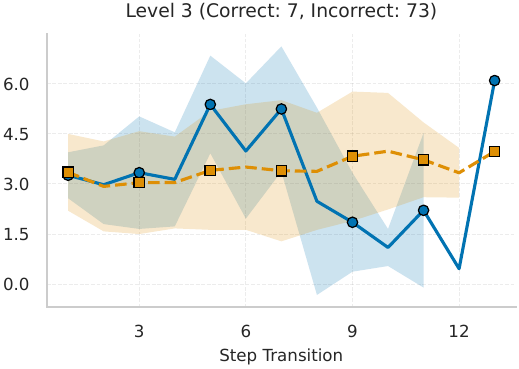}
    }
    \label{fig:aime_mathstral_csd}%
  }

  \vspace{6pt}

  % Row 2: Math-500 (5 panels)
  \subfloat[Math-500 -- MathStral-7B (KL Divergence) across difficulty levels.]{%
    \makebox[0.99\textwidth][c]{%
      \includegraphics[width=0.19\textwidth]{figs/math500_mathstral_csd_ylabel_KL_Divergence_level_1.pdf}
      \includegraphics[width=0.19\textwidth]{figs/math500_mathstral_csd_KL_Divergence_level_2.pdf}
      \includegraphics[width=0.19\textwidth]{figs/math500_mathstral_csd_KL_Divergence_level_3.pdf}
      \includegraphics[width=0.19\textwidth]{figs/math500_mathstral_csd_KL_Divergence_level_4.pdf}
      \includegraphics[width=0.19\textwidth]{figs/math500_mathstral_csd_KL_Divergence_level_5.pdf}
    }
    \label{fig:math500_mathstral_csd}%
  }%

  \vspace{6pt}

  % Row 3: MedQA (2 panels)
  \subfloat[MedQA -- Qwen-Medicine-7B (KL Divergence) across difficulty levels.]{%
    \makebox[0.99\textwidth][c]{%
      \includegraphics[width=0.29\textwidth]{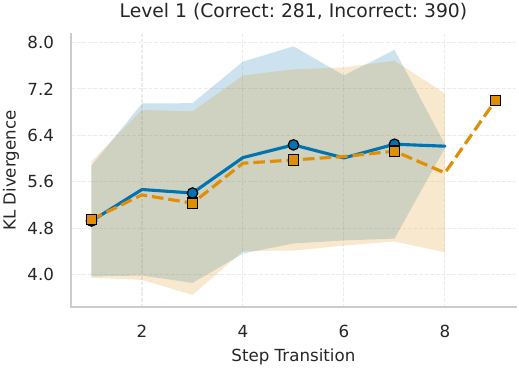}
      \includegraphics[width=0.29\textwidth]{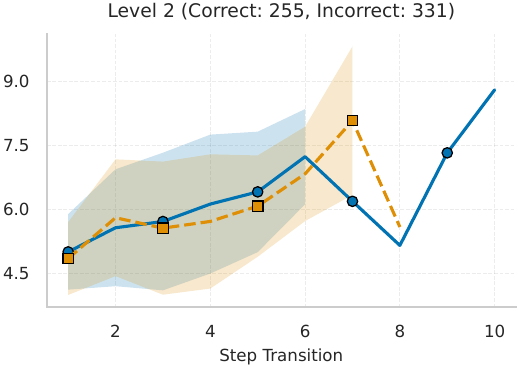}
    }
    \label{fig:medqa_csd}%
  }%

  \caption{Cross-domain CSD analysis using KL Divergence across difficulty
  levels. (a)~AIME with MathStral-7B, (b)~Math-500 with MathStral-7B, and
  (c)~MedQA with Qwen-Medicine-7B.}
  \label{fig:cross_model_domain}
\end{figure*}

To assess whether CSD patterns generalize across domains and models, 
Figure~\ref{fig:cross_model_domain} presents three dataset–model pairs: 
AIME with MathStral-7B, Math-500 with MathStral-7B, and MedQA with 
Qwen-Medicine-7B, complementing earlier analyses 
(Figs.~\ref{fig:aime_csd_levels}, \ref{fig:csd_sfc_comparison}).

Mathematical datasets show progressive volatility increase as difficulty rises. Math-500 (Fig.~\ref{fig:cross_model_domain}b) demonstrates this across five levels: Level 1 exhibits stable trajectories while Levels 4--5 display substantial fluctuations. Despite increasing volatility and varying separation across individual difficulty levels, incorrect reasoning generally displays greater instability with wider confidence bands.

Mathematical reasoning (AIME, Math-500) exhibits varying but persistent patterns distinguishing correct from incorrect trajectories, with separation strength depending on difficulty level and problem structure. Medical reasoning (MedQA, Fig.~\ref{fig:cross_model_domain}c) displays parallel, nearly identical trajectories with minimal separation, indicating 
fundamentally different local coherence dynamics.
This cross-domain analysis establishes that mathematical and medical reasoning follow fundamentally different dynamics. Mathematical reasoning 
shows difficulty-dependent patterns where local coherence provides a signal for correctness discrimination. Medical reasoning exhibits patterns where local step-level coherence does not distinguish correct from incorrect conclusions, indicating that diagnostic reasoning relies on mechanisms less reflected in step-level probability distributions.
This domain-specific finding aligns with recent work demonstrating that chain-of-thought reasoning systematically fails in clinical contexts (86.3\% of models show degradation), attributed to the complexity and fragmentation of clinical documentation\cite{wuWhyChainThought2025}. Our divergence-based analysis provides quantitative evidence for this distinction, suggesting that diagnostic reasoning may require alternative evaluation frameworks tailored to clinical inference dynamics.

\subsection{Predictive Modeling of Reasoning Reliability}
\label{subsec:predictive_modeling}

Following pattern recognition analysis, which revealed systematic differences in CSD and SFC trajectories between correct and incorrect reasoning, we now evaluate whether these quantified features can predict answer correctness. 
We employ two complementary approaches: classical machine learning models (LR, SVM, XGBoost) trained on aggregated statistical features, and sequential neural models (NN, GRU, LSTM) that process stepwise divergence sequences. 
Performance is evaluated using F1-Score (primary metric), ROC-AUC, Accuracy, and Balanced Accuracy for both CSD and SFC algorithms to enable direct algorithmic comparison.

\subsubsection{Classical Machine Learning Models}
\begin{figure*}[t]
\centering

\subfloat[F1-Score]{%
    \includegraphics[width=0.32\textwidth]{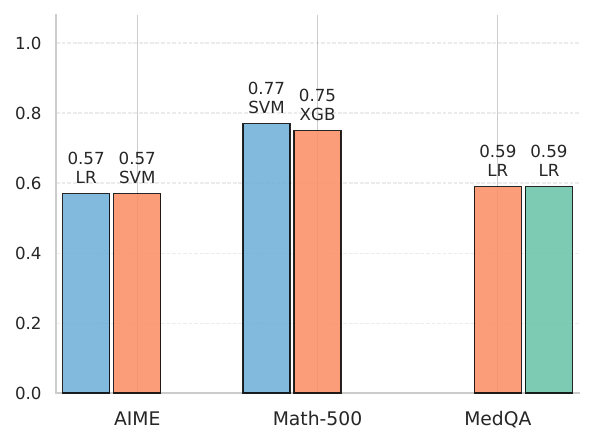}%
    \label{fig:ml_csd_f1}
}
\hfill
\subfloat[ROC-AUC]{%
    \includegraphics[width=0.32\textwidth]{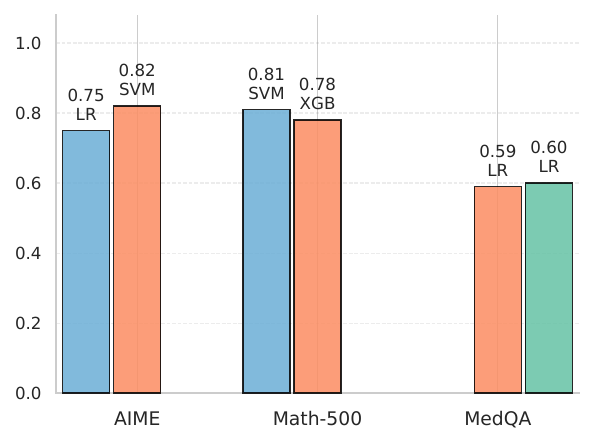}%
    \label{fig:ml_csd_roc}
}
\hfill
\subfloat[Accuracy]{%
    \includegraphics[width=0.32\textwidth]{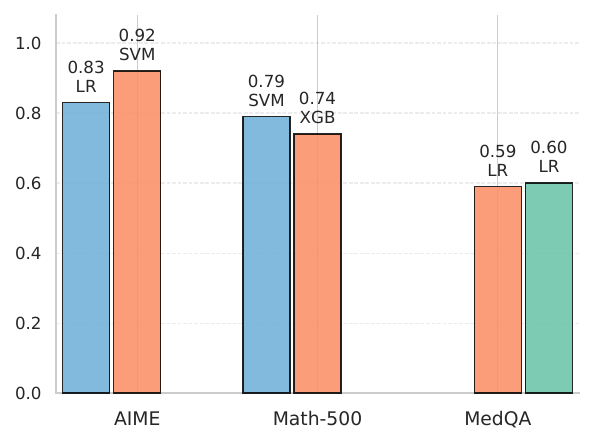}%
    \label{fig:ml_csd_acc}
}
% Shared legend below all subplots
% TikZ legend
\vspace{1mm}
\begin{center}
{\footnotesize
\begin{tikzpicture}
  % Legend box with border
  \node[draw=gray, inner sep=3pt, rounded corners=2pt] {
    \begin{tabular}{@{}c@{\hspace{8pt}}c@{\hspace{8pt}}c@{}}
      \tikz \fill[ieeesoftblue] (0,0) rectangle (0.2,0.2); MathStral &
      \tikz \fill[ieeecoral] (0,0) rectangle (0.2,0.2); Qwen2.5 &
      \tikz \fill[ieeegreen] (0,0) rectangle (0.2,0.2); Qwen\_medicine
    \end{tabular}
  };
\end{tikzpicture}
}
\end{center}
\caption{Best classical ML model performance using CSD features across datasets and LLM models showing (a) F1-Score, (b) ROC-AUC, and (c) accuracy. Each bar represents the optimal classifier (LR, SVM, or XGB) for that configuration.}
\label{fig:ml_performance_csd_metrics}
\end{figure*}
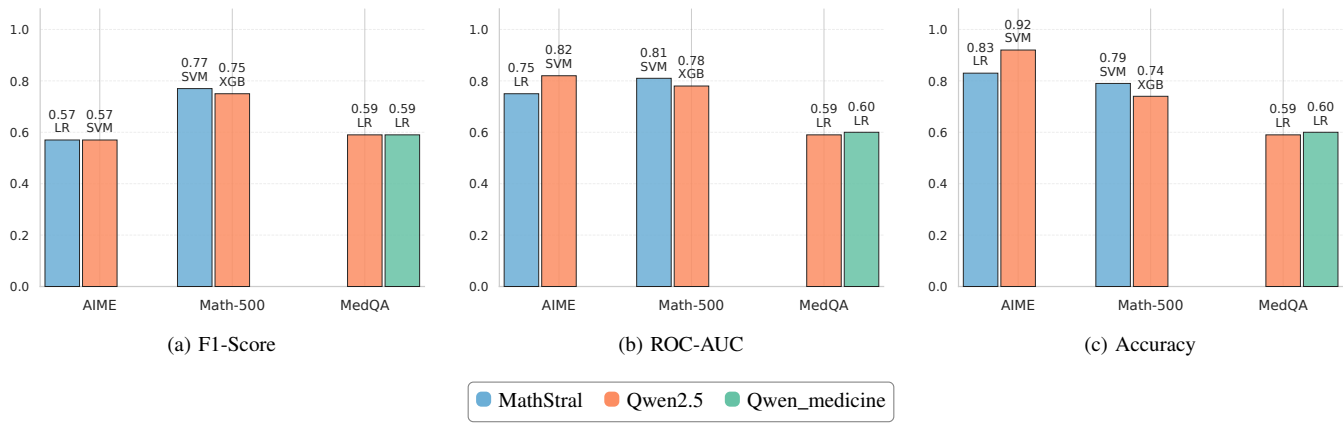

Table~\ref{tab:ml_performance_csd_sfc} presents the best-performing classifier per dataset-LLM-algorithm combination, with CSD and SFC results shown side-by-side. Figure~\ref{fig:ml_performance_csd_metrics} visualizes CSD performance across metrics. CSD consistently outperforms SFC, with the largest margin reaching +17 F1 points on AIME-Qwen2.5. SFC achieves competitive performance only on balanced mathematical settings (Math-500), indicating that local coherence provides stronger correctness discrimination than global convergence, particularly on challenging or imbalanced datasets.
The best performance on mathematical reasoning reaches F1=0.78 and ROC-AUC=0.82, while medical reasoning shows minimal signal. Detailed best SFC performance and the hyperparameters are provided in Supplementary Material Tables S2.

\input{tables/ml_performance_csd_sfc}

\subsubsection{Sequential Neural Models}

\begin{figure*}
    \centering
\subfloat[F1-Score]{%
    \includegraphics[width=0.30\textwidth]{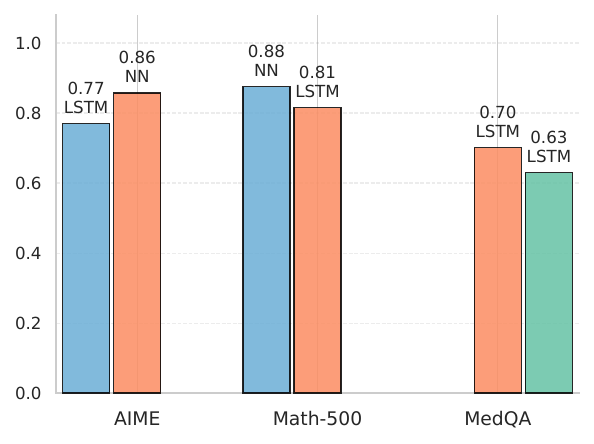}%
    \label{fig:Sequential_model_csd_f1}
}
\hfill
\subfloat[ROC-AUC]{%
    \includegraphics[width=0.30\textwidth]{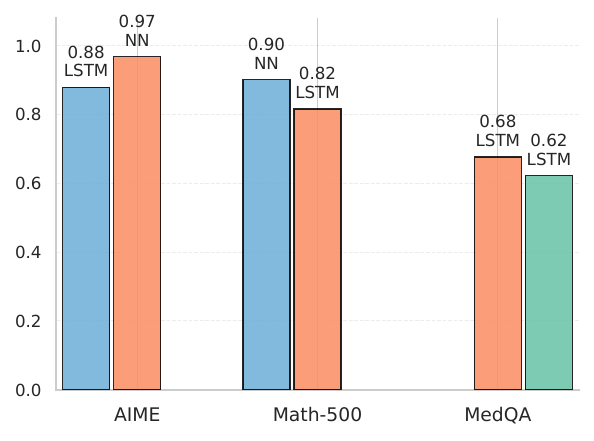}%
    \label{fig:/sequential_model_csd_roc_auc}
}
\hfill
\subfloat[Accuracy]{%
    \includegraphics[width=0.30\textwidth]{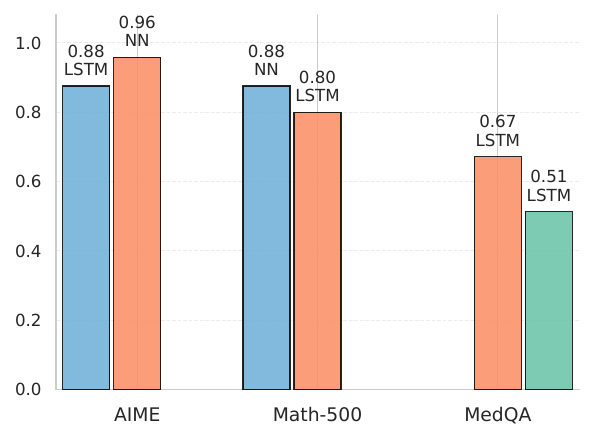}%
    \label{fig:/sequential_model__csd_accuracy}
}
% Shared legend below all subplots
% TikZ legend

\vspace{1mm}
\begin{center}
{\footnotesize
\begin{tikzpicture}
  % Legend box with border
  \node[draw=gray, inner sep=3pt, rounded corners=2pt] {
    \begin{tabular}{@{}c@{\hspace{8pt}}c@{\hspace{8pt}}c@{}}
      \tikz \fill[ieeesoftblue] (0,0) rectangle (0.25,0.25); MathStral &
      \tikz \fill[ieeecoral] (0,0) rectangle (0.25,0.25); Qwen2.5 &
      \tikz \fill[ieeegreen] (0,0) rectangle (0.25,0.25); Qwen\_medicine
    \end{tabular}
  };
\end{tikzpicture}
}
\end{center}
    \caption{Best sequential model performance using CSD features across datasets and LLM models, showing (a) F1 Score, (b) ROC-AUC, (c) Accuracy. Each bar represents the optimal architecture (NN, GRU, or LSTM) for that configuration.}
\label{fig:sequential_csd_best_performance}
\end{figure*}

While classical machine learning models establish strong baselines, their reliance on aggregated statistics prevents them from capturing temporal reasoning dynamics. We now investigate sequential neural architectures (NN, GRU, LSTM) that directly model stepwise divergence evolution. Table~\ref{tab:sequential_csd_sfc_comparison} presents sequential architecture performance across datasets. Figure~\ref{fig:sequential_csd_best_performance} visualizes CSD results across metrics.

Sequential models outperform classical models (Supplementary Material Table S3), with average F1 improvements of +13.5 points. Math-500 improves from F1=0.77 to 0.88 (+11 points), while AIME shows dramatic gains from F1=0.57 to 0.86 (+29 points). Larger improvements on challenging datasets indicate that temporal modeling becomes especially valuable when aggregated features provide a limited signal.

CSD vs SFC Performance: Sequential architectures narrow the algorithmic gap observed in classical ML. CSD still outperforms SFC in most configurations, but margins are smaller. In some cases, SFC becomes superior: on AIME-MathStral, SFC (F1 = 0.86, AUC = 0.94) surpasses CSD (F1 = 0.77, AUC = 0.88) across all metrics, suggesting that temporal modeling enhances SFC's ability to leverage global convergence patterns. Even advanced sequential architectures show limited discrimination on medical reasoning (F1=0.63-0.70), indicating domain-specific limitations rather than model capacity constraints. Detailed hyperparameters are in Supplementary Material Tables S3.

\input{tables/sequential_performance_csd_sfc}

\subsubsection{Comparison with Existing Methods}
\label{subsubsec:comparison}

Table~\ref{tab:comparison_performance} compares EvalQReason with automated reasoning evaluation methods on mathematical datasets. Embedding-based approaches (ROSCOE~\cite{golovneva2022roscoe}) achieve F1=48.2-51.6~\cite{xia2025evaluating}, while supervised methods reach F1=62.6-79.6~\cite{wang2024math,zheng2025processbench,xia2025evaluating}. EvalQReason achieves F1=88.0 without training labels, outperforming the previous best method (ReasonEval, F1=79.6) by 8.4 points. This demonstrates that probability-based divergence analysis provides stronger correctness discrimination than supervised approaches while eliminating annotation costs.

\input{tables/comparing_performance}

\section{Limitations and Future Work}
\label{sec:limitations}

EvalQReason provides a process-level framework for analyzing reasoning behavior in LLMs, with several directions for future work. Our experiments focus on open-source 7B-parameter models to enable direct access to token-level logits, extending to larger models would clarify whether reasoning stability improves with scale. 

An important future direction involves applying EvalQReason to 
adversarial and safety-critical scenarios to investigate whether unsafe inputs induce distinctive instability patterns, informing process-level detection mechanisms and reasoning-aware guardrails.

Finally, our cross-domain analysis reveals that divergence-based 
evaluation provides strong signals for mathematical reasoning but 
limited discrimination for medical diagnostic tasks. This suggests that clinical reasoning may require complementary evaluation approaches tailored to diagnostic inference dynamics, motivating research on domain-adaptive frameworks beyond step-level probability analysis.

\section{Conclusion}
\label{sec:conclusion}
This study introduced EvalQReason, a framework for quantifying LLM reasoning behavior through step-level divergence analysis using CSD and SFC algorithms. Experiments across mathematical and medical domains reveal that correct reasoning exhibits stable, low-divergence trajectories, while incorrect reasoning shows volatility that intensifies with task complexity.
CSD consistently outperforms SFC for correctness prediction, achieving up to 17 F1-point improvements in classical ML. Sequential models that capture temporal reasoning patterns deliver further gains (+6 to +29 F1 points), reaching F1=0.88 and ROC-AUC=0.97 on mathematical tasks. Critically, reasoning 
dynamics prove domain-specific: mathematical reasoning exhibits clear divergence-based patterns, while medical reasoning shows minimal discriminative signal, indicating fundamentally different inference mechanisms.

EvalQReason enables scalable and interpretable evaluation of LLM reliability by shifting assessment from final outcomes to process-level reasoning dynamics, addressing a critical gap in trustworthy AI assessment. This divergence-based perspective lays the groundwork for extending process-level reasoning analysis to adversarial robustness and safety monitoring, where reasoning-pattern analysis may support early detection of harmful or manipulative inputs—an essential capability for deploying LLMs in high-stakes applications.

\appendix
\section{Supplementary Materials}
\label{sec:Supplementary}

\section*{Contents}

\begin{itemize}
\item \textbf{Section S1: Prompts for Reasoning Chain Generation}\\
Complete prompts used for mathematical and medical reasoning (Figure S1)

\item \textbf{Section S2: Statistical Metrics for Divergence Analysis}\\
Mathematical definitions of KL divergence, JS divergence, Hellinger distance, cosine similarity, and entropy

\item \textbf{Section S3: Additional Performance Visualizations}\\
Performance comparisons between classical ML and sequential models, SFC performance plots, and cross-domain CSD analysis (Figures S2--SX)

\item \textbf{Section S4: Hyperparameter Configurations}\\
Complete hyperparameters for classical ML and sequential models (Tables S1--S4)

\end{itemize}

\subsection{S1. Prompts for Reasoning Chain Generation}

Fig.~1 presents the complete prompts used to elicit structured reasoning chains from LLMs across mathematical and medical domains. Specifically, the mathematical prompt instructs models to act as expert mathematicians, solving problems through step-by-step reasoning, while the medical prompt frames the model as a medical board examination tutor, guiding students through USMLE-style multiple-choice questions using structured clinical reasoning.

\begin{figure}[b]
  \centering
\includegraphics[width=\linewidth]{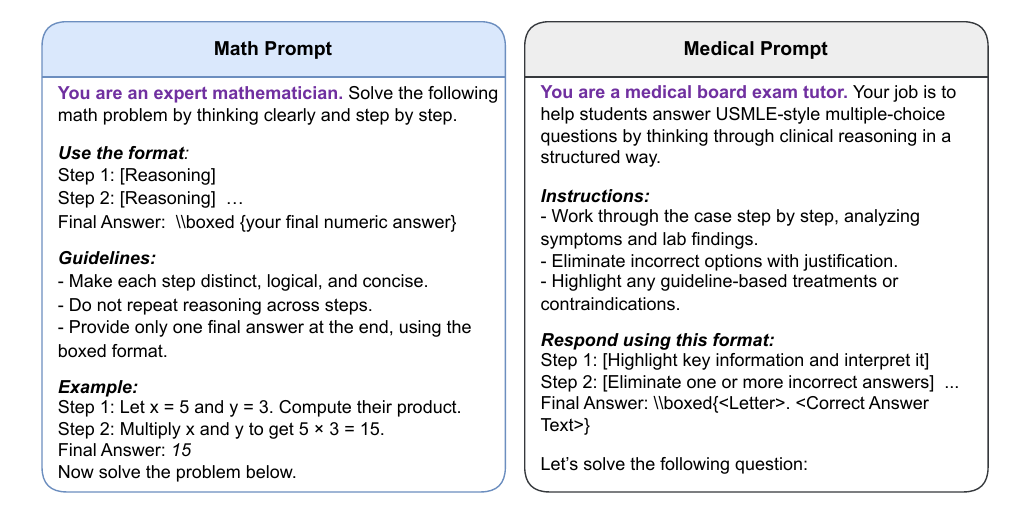}
  \caption{Math and medical prompts are used to elicit structured step-by-step reasoning from LLMs.}
  \label{fig:prompts}
\end{figure}

\subsection{S2. Statistical Metrics for Divergence Analysis}

EvalQReason employs five complementary statistical metrics to quantify distributional relationships in CSD and SFC algorithms. These metrics capture distinct aspects of reasoning dynamics through divergence measures, similarity measures, and uncertainty quantification.

\subsubsection{S2.1 Divergence Measures}

\textbf{Kullback-Leibler (KL) Divergence.} Quantifies information loss when Q approximates P, providing a directed measure of distributional difference:
$$KL(P \| Q) = \sum_{x} P(x) \log \frac{P(x)}{Q(x)}$$

KL divergence is non-symmetric and unbounded, making it suitable for measuring directed distributional shifts between consecutive reasoning steps.

\textbf{Jensen-Shannon (JS) Divergence.} A symmetric and bounded divergence measure:
$$JSD(P \| Q) = \frac{1}{2}KL(P \| M) + \frac{1}{2}KL(Q \| M)$$
where $M = \frac{1}{2}(P + Q)$. JSD is symmetric and bounded in [0, 1], making it suitable for comparing reasoning steps without directional bias.

\textbf{Hellinger Distance.} A bounded metric sensitive to shifts in probability mass:
$$H(P, Q) = \frac{1}{\sqrt{2}} \left\| \sqrt{P} - \sqrt{Q} \right\|$$

Hellinger distance is particularly sensitive to transitions in high-probability regions, highlighting major distributional shifts.

\subsubsection{S2.2 Similarity Measure}

\textbf{Cosine Similarity.} Measures angular alignment between probability distributions:
$$\cos(P, Q) = \frac{P \cdot Q}{\|P\| \|Q\|}$$

Values range from -1 (opposite directions) to +1 (identical directions). Higher values indicate that consecutive reasoning steps maintain consistent distributional patterns.

\subsubsection{S2.3 Uncertainty Measure}

\textbf{Entropy Difference.} Measures changes in uncertainty between reasoning steps. The entropy of distribution P is:
$$H(P) = -\sum_{x} P(x) \log P(x)$$

For CSD, we compute signed entropy change: $\Delta H_i = H(P_{i+1}) - H(P_i)$. 
For SFC, we compute absolute deviation: $\Delta H_i = |H(P_i) - H(P_{final})|$.

\subsubsection{S2.4 Implementation Details}

All divergences are computed over the full vocabulary without truncation, using natural logarithms (units: nats). For numerical stability, we apply additive smoothing: $\tilde{p}_i = \bar{p}_i + \epsilon$ where $\epsilon = 10^{-7}$.

\subsection{S3. Additional Performance Visualizations}

This section provides supplementary visualizations supporting the analysis in the main text, including comparative performance analysis and step-to-final convergence (SFC) patterns.

\subsubsection{S3.1. Classical ML vs. Sequential Model Comparison}

Table~S presents a comprehensive comparison of predictive performance between classical machine learning and sequential neural network models across all datasets and algorithms. Sequential models consistently outperform classical approaches, demonstrating the value of temporal modeling in capturing reasoning dynamics. Performance gains are particularly pronounced on mathematical datasets, where sequential models improve F1-scores by 6--29 points over classical methods, with the largest improvements observed on AIME. This superior performance reflects sequential models' ability to capture temporal patterns such as mid-step instability and progressive degradation that static feature aggregation cannot detect.

\input{tables/ml_sequential_comparison}

\subsubsection{S3.2. SFC Performance Analysis}

To complement the CSD performance visualizations in the main text (Figures~2 and~3), this subsection provides corresponding performance plots for step-to-final convergence (SFC) features.

Figures~S2 present SFC-based performance to complement the CSD visualizations in the main text. SFC achieves competitive performance on Math-500 but underperforms CSD on AIME.

\begin{figure*}[t]
\centering

\subfloat[F1-Score]{%
    \includegraphics[width=0.32\textwidth]{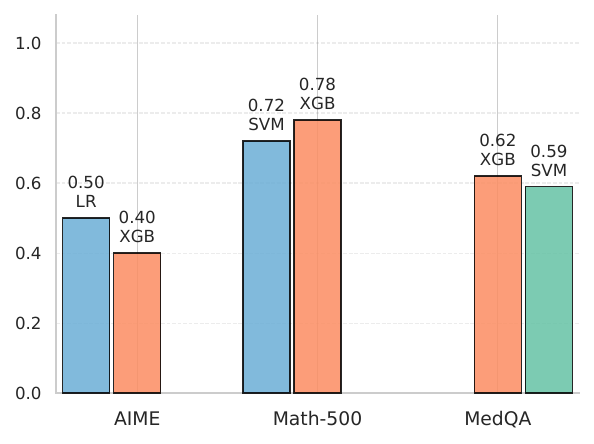}%
    \label{fig:ml_sfc_f1}
}
\hfill
\subfloat[ROC-AUC]{%
    \includegraphics[width=0.32\textwidth]{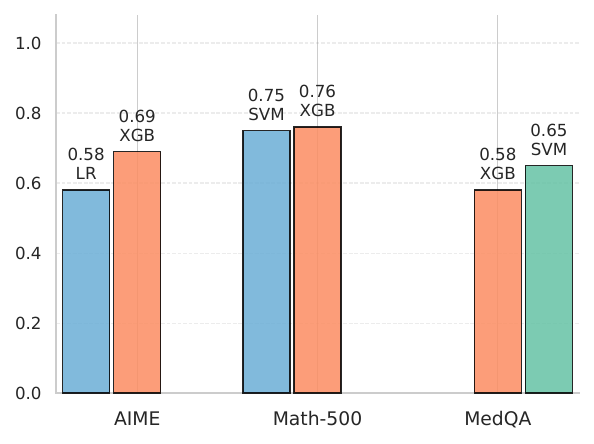}%
    \label{fig:ml_sfc_roc}
}
\hfill
\subfloat[Accuracy]{%
    \includegraphics[width=0.32\textwidth]{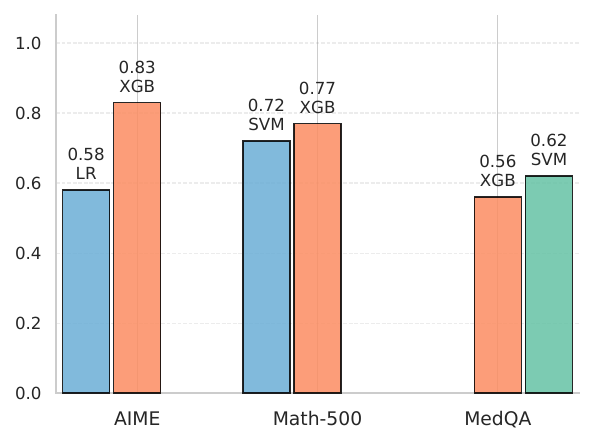}%
    \label{fig:ml_sfc_acc}
}
% Shared legend below all subplots
% TikZ legend
\vspace{2mm}
\begin{center}
{\footnotesize
\begin{tikzpicture}
  % Legend box with border
  \node[draw=gray, inner sep=4pt, rounded corners=2pt] {
    \begin{tabular}{@{}c@{\hspace{8pt}}c@{\hspace{8pt}}c@{}}
      \tikz \fill[ieeesoftblue] (0,0) rectangle (0.2,0.2); MathStral &
      \tikz \fill[ieeecoral] (0,0) rectangle (0.2,0.2); Qwen2.5 &
      \tikz \fill[ieeegreen] (0,0) rectangle (0.2,0.2); Qwen\_medicine
    \end{tabular}
  };
\end{tikzpicture}
}
\end{center}
\caption{Best Classical ML model performance using SFC features across datasets and LLM models showing (a) F1-Score, (b) ROC-AUC, and (c) Accuracy. Each bar 
represents the optimal classifier (LR, SVM, or XGB) for that configuration.}
\label{fig:ml_performance_sfc_metrics}
\end{figure*}
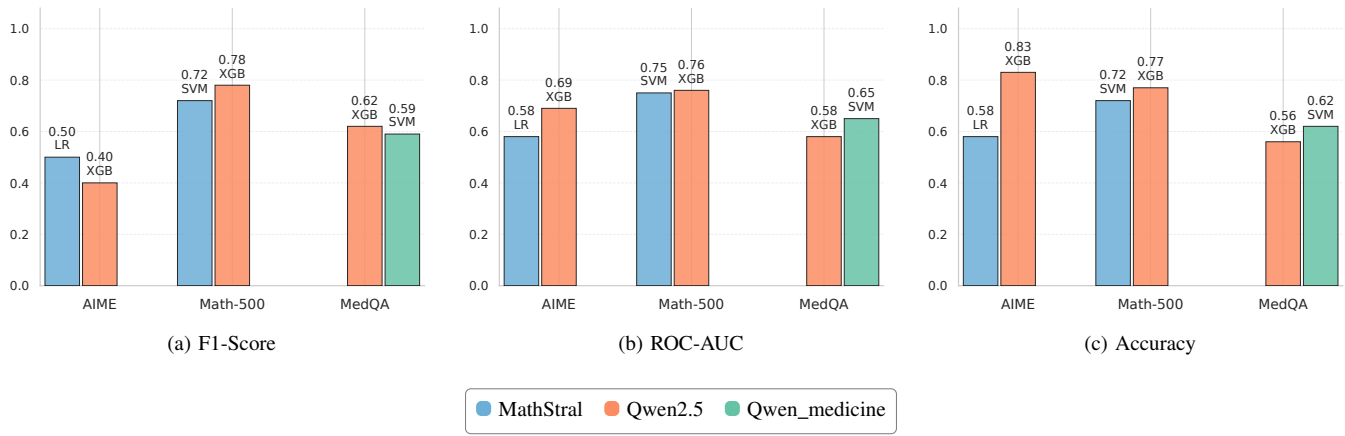

Figures~S3 presents the best-performing 
sequential model results using SFC features. Sequential architectures 
substantially narrow the CSD-SFC performance gap observed in classical models, 
with SFC achieving competitive or superior results on several configurations.

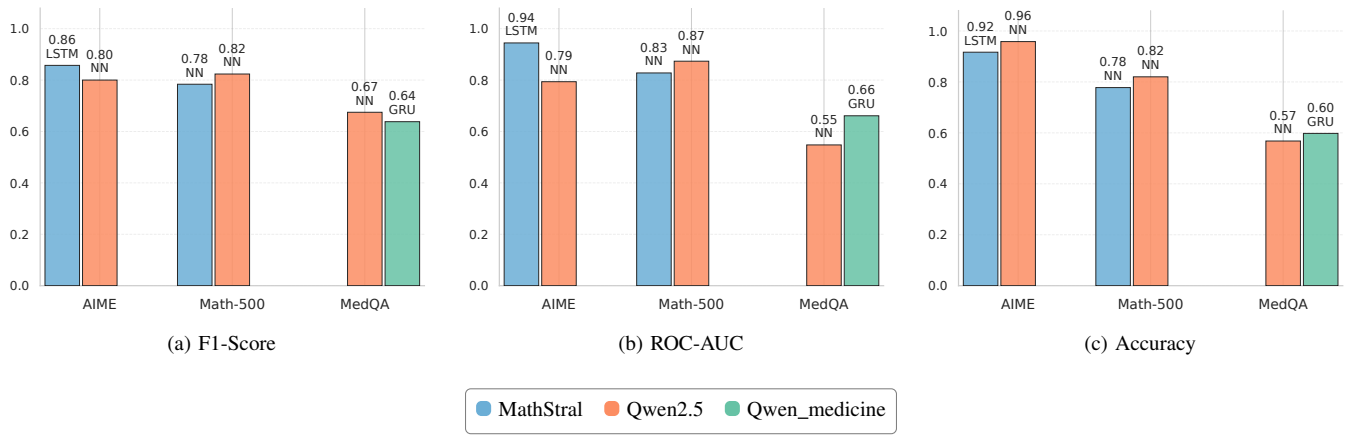
\begin{figure*}
    \centering

\subfloat[F1-Score]{%
    \includegraphics[width=0.32\textwidth]{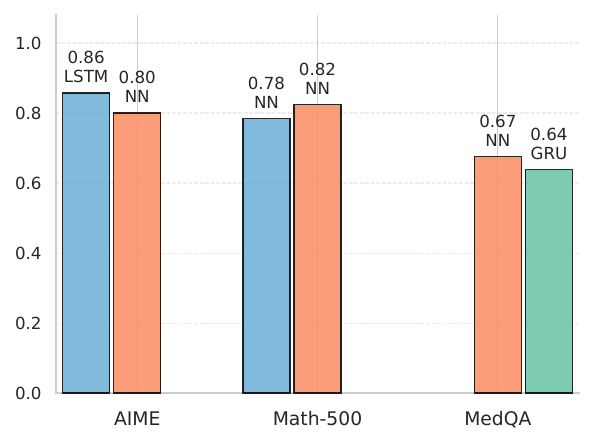}%
    \label{fig:Sequential_model_sfc_f1}
}
\hfill
\subfloat[ROC-AUC]{%
    \includegraphics[width=0.32\textwidth]{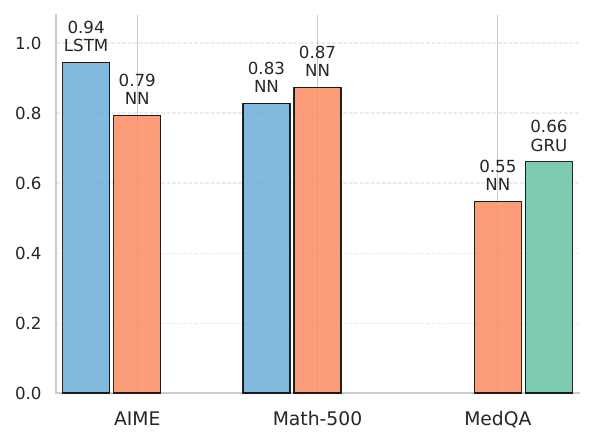}%
    \label{fig:/sequential_model_sfc_roc_auc}
}
\hfill
\subfloat[Accuracy]{%
    \includegraphics[width=0.32\textwidth]{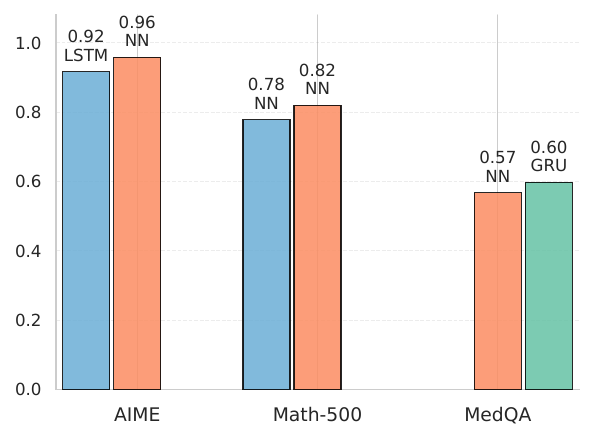}%
    \label{fig:/sequential_model_sfc_accuracy}
}
% Shared legend below all subplots
% TikZ legend
\vspace{2mm}
\begin{center}
{\footnotesize
\begin{tikzpicture}
  % Legend box with border
  \node[draw=gray, inner sep=4pt, rounded corners=2pt] {
    \begin{tabular}{@{}c@{\hspace{8pt}}c@{\hspace{8pt}}c@{}}
      \tikz \fill[ieeesoftblue] (0,0) rectangle (0.2,0.2); MathStral &
      \tikz \fill[ieeecoral] (0,0) rectangle (0.2,0.2); Qwen2.5 &
      \tikz \fill[ieeegreen] (0,0) rectangle (0.2,0.2); Qwen\_medicine
    \end{tabular}
  };
\end{tikzpicture}
}
\end{center}
    
    \caption{Best Sequential model performance using SFC features across datasets and LLM models, showing (a) F1-Score, (b) ROC-AUC, and (c) Accuracy. Each bar represents the optimal architecture (NN, GRU, or LSTM) for that configuration.}
\label{fig:sequential_sfc_best_performance}
\end{figure*}

\subsection{S4. Hyperparameter Configurations}

This section provides detailed hyperparameter configurations for all predictive models to ensure full reproducibility of results reported in the main paper.

\subsubsection{S4.1. Classical Machine Learning Models}

Table~S2 reports hyperparameters for the best-performing classical ML classifiers using both CSD and SFC features. All models were optimized through grid search, randomized search, and Bayesian optimization, with final configurations selected based on F1-Score on validation sets.

\input{tables/ml_hyperparameter_csd_sfc}

\subsubsection{S4.2. Sequential Neural Network Models}

Table~S3 reports hyperparameters for the best-performing sequential models (NN, GRU, LSTM) using both CSD and SFC features. Sequential models were optimized by exploring learning rate, L2 regularization, and dropout configurations. All models use hidden dimensions of 64 and 128, ReLU activation, batch size 32, and early stopping with patience of 50 epochs. Final configurations were selected based on F1-Score on validation sets.

\input{tables/seq_hyperparameter_csd_sfc}

\bibliographystyle{IEEEtran}
\bibliography{references}

\begin{IEEEbiographynophoto}{Shaima Ahmad Freja}
received the M.Sc. degree in data science from the University of Stavanger, Norway. She is currently pursuing the Ph.D. degree in trustworthy artificial intelligence with the University of Stavanger. Her research interests include the security, privacy, and reliability of generative AI systems.
\end{IEEEbiographynophoto}

\begin{IEEEbiographynophoto}{Ferhat Ozgur Catak} received the Ph.D. degree in informatics from Istanbul University, Istanbul, Turkey, in 2014. He is currently an Associate Professor with the Department of Electrical Engineering and Computer Science, University of Stavanger, Stavanger, Norway. He has previously held research positions at the Norwegian University of Science and Technology (NTNU) and Simula Research Laboratory, Norway. His research interests include trustworthy and robust artificial intelligence, secure and explainable machine learning, radio frequency signal processing, spectrum sensing, and next-generation (6G) wireless communication systems. He has authored numerous peer-reviewed journal and conference papers and a research monograph on trustworthy AI. Dr. Catak is a Senior Member of the IEEE and currently serves as the Chair of the IEEE Norway Communications Society Chapter.
\end{IEEEbiographynophoto}

\begin{IEEEbiographynophoto}{Betul Yurdem} received the B.Sc. and M.Sc. degrees in Electrical and Electronics Engineering from Dokuz Eylul University, Turkiye. She is currently a Research Assistant and Ph.D. candidate at Izmir Bakircay University. Her research interests include trustworthy AI, vision-language models, and biomedical signal processing.
\end{IEEEbiographynophoto}

\begin{IEEEbiographynophoto}{Chunming Rong}
is a Professor and the head of the Data-centered and Secure Computing (DSComputing) at the University of Stavanger (UiS) and Chief Scientist (adjunct) at NORCE Norwegian Research Centre. His research work focuses on cloud computing, data analytics, cyber security, and blockchain. He served as a co-chair of IEEE Blockchain in 2018, as the chair of IEEE Cloud Computing from 2017 to 2020, and has extensive experience in managing large-scale R\&D projects, both in Norway and the EU. He is a Senior Member of the IEEE.
\end{IEEEbiographynophoto}

\end{document}

%% file: tables/datasets.tex
\begin{table}[t]
  \renewcommand{\arraystretch}{1}   % tighter rows
  \setlength{\tabcolsep}{6pt}         % tighter cols

  \centering
  \caption{Overview of Datasets and LLMs Used for Reasoning Generation.}
  \label{tab:datasets}
  % l l r l X = last column auto-wraps
  \begin{tabularx}{\linewidth}{l l r l >{\raggedright\arraybackslash}X}
    \toprule
    \textbf{Dataset} & \textbf{Domain} & \textbf{Size} & \textbf{Difficulty} & \textbf{LLMs Used} \\
    \midrule
    Math-AIME & Math QA & 240  & 3 Levels & \makecell[l]{Qwen2.5-7B,\\ MathStral-7B} \\[8pt]
    \midrule
    Math-500  & Math QA & 500  & 5 Levels & \makecell[l]{Qwen2.5-7B,\\ MathStral-7B} \\[8pt]
    \midrule
    Med-QA    & Med MCQ & 1{,}243 & 2 Levels & \makecell[l]{Qwen2.5-7B,\\ Qwen-Medicine-7B} \\
    \bottomrule
  \end{tabularx}
\end{table}

%% file: tables/stage1_dataset.tex
\begin{table}[t]
  \renewcommand{\arraystretch}{1}   % tighter rows
  \setlength{\tabcolsep}{2.5pt}       % slightly narrower columns

  \centering
  \caption{Curated Reasoning Dataset Statistics for Stage 1, Including Correctness and Accuracy of Each Model.}
  \label{tab:stage1_dataset}

  %\begin{tabularx}{\linewidth}{l r l r r r}
  \begin{tabular}{l r l r r r}
    \toprule
    \textbf{Dataset} & \textbf{Size} & \textbf{LLMs} & \textbf{Correct} & \textbf{In-correct} & \textbf{Accuracy (\%)} \\
    \midrule
    Math-AIME & 240 & \makecell[l]{MathStral-7B\\Qwen2.5-7B} & \makecell[r]{62\\33} & \makecell[r]{178\\201} & \makecell[r]{25.8\\14.1} \\[8pt]
    \midrule
    Math-500  & 500 & \makecell[l]{MathStral-7B\\Qwen2.5-7B} & \makecell[r]{207\\239} & \makecell[r]{270\\256} & \makecell[r]{43.4\\48.3} \\[8pt]
    \midrule
    Med-QA    & 1{,}243 & \makecell[l]{Qwen2.5-7B\\Qwen-Med-7B} & \makecell[r]{592\\536} & \makecell[r]{651\\721} & \makecell[r]{47.6\\42.6} \\
    \bottomrule
  \end{tabular}
\end{table}

%% file: tables/step_group_results.tex
\begin{table}[t]
\centering
\caption{Accuracy (\%) across datasets, Models, and reasoning step-length groups.
Cell shading indicates relative accuracy within the table (darker = higher).}
\label{tab:step_group_results}
\setlength{\tabcolsep}{4pt}
\renewcommand{\arraystretch}{1.1}

\begin{tabular}{llc|ccc}
\toprule
\textbf{Dataset} & \textbf{Model} & \textbf{Thresholds} & \textbf{Short} & \textbf{Medium} & \textbf{Long} \\
\midrule
\multirow{2}{*}{Math-AIME}
 & MathStral-7B & 7--11
 & \cellcolor{ieeeblue!35}34.69
 & \cellcolor{ieeeblue!20}20.83
 & \cellcolor{ieeeblue!10}13.64 \\
 & Qwen2.5-7B & 6--10
 & \cellcolor{ieeeblue!20}19.57
 & \cellcolor{ieeeblue!12}13.77
 & \cellcolor{ieeeblue!5}10.00 \\
\midrule
\multirow{2}{*}{Math-500}
 & MathStral-7B & 4--7
 & \cellcolor{ieeeblue!45}49.15
 & \cellcolor{ieeeblue!30}40.85
 & \cellcolor{ieeeblue!10}16.67 \\
 & Qwen2.5-7B & 6--10
 & \cellcolor{ieeeblue!55}59.81
 & \cellcolor{ieeeblue!30}41.10
 & \cellcolor{ieeeblue!20}31.11 \\
\midrule
\multirow{2}{*}{Med-QA}
 & Qwen-7B & 3--5
 & \cellcolor{ieeeblue!45}51.13
 & \cellcolor{ieeeblue!35}45.56
 & \cellcolor{ieeeblue!55}58.21 \\
 & Qwen-Med-7B & 3--5
 & \cellcolor{ieeeblue!30}39.17
 & \cellcolor{ieeeblue!35}45.99
 & \cellcolor{ieeeblue!30}44.38 \\
\bottomrule
\end{tabular}

\vspace{1.5mm}
\footnotesize
\textbf{Accuracy scale (approx.):}
\colorbox{ieeeblue!5}{\phantom{XX}}$\le$10\%
\quad
\colorbox{ieeeblue!15}{\phantom{XX}}15--20\%
\quad
\colorbox{ieeeblue!30}{\phantom{XX}}30--40\%
\quad
\colorbox{ieeeblue!45}{\phantom{XX}}40--50\%
\quad
\colorbox{ieeeblue!55}{\phantom{XX}}$\ge$55\%
\end{table}

%% file: tables/ml_performance_csd_sfc.tex
\begin{table*}[t]
\centering
\setlength{\tabcolsep}{5pt}

\caption{Performance of LR, SVM, and XGB ML models across dataset-LLM pairs using CSD and SFC features.}
\label{tab:ml_performance_csd_sfc}

\makebox[\textwidth]{%
\begin{tabular}{l l l 
                c c c c 
                c c c c}
\toprule
\multirow{2}{*}{\textbf{Dataset}} & 
\multirow{2}{*}{\textbf{LLM Model}} & 
\multirow{2}{*}{\textbf{ML model}} & 
\multicolumn{4}{c}{\textbf{CSD}} & 
\multicolumn{4}{c}{\textbf{SFC}} \\
\cmidrule(lr){4-7} \cmidrule(lr){8-11}
& & & \textbf{F1} & \textbf{AUC} & \textbf{Acc} & \textbf{BalAcc} & \textbf{F1} & \textbf{AUC} & \textbf{Acc} & \textbf{BalAcc} \\
\midrule

% ----------------- math_500 -----------------

\multirow{6}{*}{Math-500}
  & \multirow{3}{*}{MathStral-7B}
    & LR  & 0.76 & 0.82 & 0.79 & 0.79 & 0.67 & 0.74 & 0.73 & 0.72 \\
  & & \textbf{SVM} & \textbf{0.77} & \textbf{0.81} & \textbf{0.79} & \textbf{0.79}
                  & \textbf{0.72} & \textbf{0.75} & \textbf{0.72} & \textbf{0.74} \\
  & & XGB & 0.74 & 0.78 & 0.77 & 0.77 & 0.68 & 0.78 & 0.71 & 0.71 \\

\cmidrule(lr){2-11}

  & \multirow{3}{*}{Qwen2.5-7B}
    & LR  & 0.71 & 0.77 & 0.72 & 0.72 & 0.72 & 0.78 & 0.72 & 0.72 \\
  & & SVM & 0.75 & 0.76 & 0.74 & 0.74 & 0.75 & 0.77 & 0.75 & 0.75 \\
  & & \textbf{XGB} & \textbf{0.75} & \textbf{0.78} & \textbf{0.74} & \textbf{0.74}
                  & \textbf{0.78} & \textbf{0.76} & \textbf{0.77} & \textbf{0.78} \\

\midrule
\addlinespace

% ----------------- math_aime -----------------

\multirow{6}{*}{AIME}
  & \multirow{3}{*}{MathStral-7B}
    & \textbf{LR} & \textbf{0.57} & \textbf{0.75} & \textbf{0.83} & \textbf{0.70}
                  & \textbf{0.50} & \textbf{0.58} & \textbf{0.58} & \textbf{0.67} \\
  & & SVM & 0.50 & 0.40 & 0.75 & 0.67 & 0.50 & 0.45 & 0.67 & 0.67 \\
  & & XGB & 0.53 & 0.60 & 0.75 & 0.69 & 0.38 & 0.55 & 0.72 & 0.59 \\

\cmidrule(lr){2-11}

  & \multirow{3}{*}{Qwen2.5-7B}
    & LR  & 0.33 & 0.56 & 0.89 & 0.60 & 0.36 & 0.61 & 0.70 & 0.65 \\
  & & \textbf{SVM} & \textbf{0.57} & \textbf{0.82} & \textbf{0.92} & \textbf{0.70}
                  & 0.32 & 0.64 & 0.72 & 0.60 \\
  & & XGB & 0.50 & 0.67 & 0.92 & 0.67 
                  & \textbf{0.40} & \textbf{0.69} & \textbf{0.83} & \textbf{0.65} \\

\midrule
\addlinespace

% ----------------- medical -----------------

\multirow{6}{*}{MedQA}
  & \multirow{3}{*}{Qwen2.5-7B}
    & \textbf{LR} & \textbf{0.59} & \textbf{0.59} & \textbf{0.59} & \textbf{0.59}
                  & 0.55 & 0.58 & 0.58 & 0.57 \\
  & & SVM & 0.58 & 0.60 & 0.60 & 0.60 & 0.56 & 0.62 & 0.60 & 0.60 \\
  & & XGB & 0.58 & 0.58 & 0.61 & 0.61 
                  & \textbf{0.62} & \textbf{0.58} & \textbf{0.56} & \textbf{0.57} \\

\cmidrule(lr){2-11} 

  & \multirow{3}{*}{Qwen-Med-7B}
    & \textbf{LR} & \textbf{0.59} & \textbf{0.60} & \textbf{0.60} & \textbf{0.61}
                  & 0.58 & 0.67 & 0.62 & 0.62 \\
  & & SVM & 0.58 & 0.63 & 0.65 & 0.64 
                  & \textbf{0.59} & \textbf{0.65} & \textbf{0.62} & \textbf{0.62} \\
  & & XGB & 0.54 & 0.60 & 0.59 & 0.59 
                  & 0.56 & 0.56 & 0.62 & 0.61 \\

\bottomrule
\end{tabular}
}
\end{table*}

%% file: tables/sequential_performance_csd_sfc.tex
\begin{table*}[t]
\centering
\caption{Performance of NN, GRU, and LSTM architectures across dataset-LLM pairs using CSD and SFC features.}
\label{tab:sequential_csd_sfc_comparison}
\begin{tabular}{llcccccccccc}
\toprule
\multirow{2}{*}{\textbf{Dataset}} &
\multirow{2}{*}{\textbf{LLM Model}} &
\multicolumn{5}{c}{\textbf{CSD}} & \multicolumn{5}{c}{\textbf{SFC}} \\
\cmidrule(lr){3-7}\cmidrule(lr){8-12}
& & \textbf{Model} & \textbf{F1} & \textbf{AUC} & \textbf{Acc} & \textbf{BalAcc} & \textbf{Model} & \textbf{F1} & \textbf{AUC} & \textbf{Acc} & \textbf{BalAcc} \\
\midrule
% -------------------------- Math-500 ---------------------------------
\multirow{6}{*}{Math-500}
& \multirow{3}{*}{MathStral-7B}
& \textbf{NN}   & \textbf{0.88} & \textbf{0.90} & \textbf{0.88} & \textbf{0.89} & \textbf{NN}   & \textbf{0.78} & \textbf{0.83} & \textbf{0.78 }& \textbf{0.80} \\
& & GRU  & 0.77 & 0.86 & 0.75 & 0.77 & LSTM & 0.75 & 0.80 & 0.72 & 0.75 \\
& & LSTM & 0.77 & 0.84 & 0.75 & 0.77 & GRU  & 0.74 & 0.85 & 0.71 & 0.74 \\
\cmidrule(lr){2-12} 
& \multirow{3}{*}{Qwen2.5-7B}
& LSTM & 0.81 & 0.82 & 0.80 & 0.81 & NN   & \textbf{0.82} & \textbf{0.87} & \textbf{0.82} & \textbf{0.82} \\
& & GRU  & 0.80 & 0.82 & 0.76 & 0.76 & LSTM & 0.80 & 0.83 & 0.80 & 0.80 \\
& & NN   & 0.80 & 0.88 & 0.76 & 0.77 & GRU  & 0.79 & 0.76 & 0.76 & 0.77 \\
\midrule
% -------------------------- AIME ---------------------------------
\multirow{6}{*}{AIME}
& \multirow{3}{*}{MathStral-7B}
& \textbf{LSTM} & \textbf{0.77} & \textbf{0.88} & \textbf{0.88} &\textbf{ 0.86} & \textbf{LSTM} & \textbf{0.86} & \textbf{0.94 }& \textbf{0.92} & \textbf{0.94} \\
& & GRU  & 0.77 & 0.80 & 0.88 & 0.86 & GRU  & 0.80 & 0.94 & 0.88 & 0.92 \\
& & NN   & 0.77 & 0.82 & 0.88 & 0.86 & NN   & 0.71 & 0.94 & 0.79 & 0.86 \\
\cmidrule(lr){2-12}
& \multirow{3}{*}{Qwen2.5-7B}
& \textbf{NN}   & \textbf{0.86} & \textbf{0.97} & \textbf{0.96} & \textbf{0.98} & NN   & 0.80 & 0.79 & 0.96 & 0.83 \\
& & GRU  & 0.80 & 1.00 & 0.96 & 0.83 & GRU  & 0.73 & 0.90 & 0.92 & 0.87 \\
& & LSTM & 0.75 & 0.90 & 0.92 & 0.95 & LSTM & 0.73 & 0.91 & 0.92 & 0.87 \\
\midrule
% -------------------------- MedQA ---------------------------------
\multirow{6}{*}{MedQA}
& \multirow{3}{*}{Qwen2.5-7B}
& \textbf{LSTM} & \textbf{0.70} & \textbf{0.68} & \textbf{0.67} & \textbf{0.68} & \textbf{NN}   & \textbf{0.67} & \textbf{0.55} &\textbf{ 0.57} & \textbf{0.58} \\
& & GRU  & 0.69 & 0.67 & 0.63 & 0.64 & GRU  & 0.67 & 0.64 & 0.53 & 0.55 \\
& & NN   & 0.69 & 0.62 & 0.61 & 0.62 & LSTM & 0.67 & 0.66 & 0.53 & 0.55 \\
\cmidrule(lr){2-12}
& \multirow{3}{*}{Qwen-Med-7B}
& \textbf{LSTM} & \textbf{0.63} & \textbf{0.62} & \textbf{0.51} & \textbf{0.57} & \textbf{GRU}  & \textbf{0.64} & \textbf{0.66} & \textbf{0.60} & \textbf{0.63} \\
& & GRU  & 0.63 & 0.63 & 0.55 & 0.59 & NN   & 0.62 & 0.59 & 0.52 & 0.58 \\
& & NN   & 0.63 & 0.63 & 0.51 & 0.56 & LSTM & 0.62 & 0.62 & 0.50 & 0.56 \\
\bottomrule
\end{tabular}
\end{table*}

%% file: tables/comparing_performance.tex
\begin{table}[t]
\centering
\caption{Performance Comparison with Reasoning Evaluation Methods on Mathematical Reasoning Tasks, Results reported on MATH-500 and MR-MATH datasets}
\label{tab:comparison_performance}
\begin{tabular}{l|c|c}
\hline
\textbf{Method} & \textbf{F1} & \textbf{AUC}  \\
\hline
\multicolumn{3}{c}{\textit{Embedding-Based}} \\
\hline
ROSCOE-SA~\cite{xia2025evaluating} & 48.2 & 57.5  \\
ROSCOE-SS~\cite{xia2025evaluating} & 51.6 & 49.6 \\
\hline
\multicolumn{3}{c}{\textit{Supervised (Require Training)}} \\
\hline
Math-Shepherd~\cite{xia2025evaluating} & 70.1 & 77.3  \\
ProcessBench (PRM-Qwen2.5-Math-7B-PRM800K)~\cite{zheng2025processbench} & 62.6 & --  \\
ProcessBench (Critic-QwQ-32B-Preview)~\cite{zheng2025processbench} & 78.7 & --  \\
ReasonEval~\cite{xia2025evaluating} & 79.6 & 90.8  \\
\hline
\multicolumn{3}{c}{\textit{Unsupervised (No Training)}} \\
\hline
\rowcolor{ieeeblue!10}
\textbf{EvalQReason-CSD (Ours)} & \textbf{88.0} & \textbf{90.0} \\
\hline
\end{tabular}
\end{table}

%% file: tables/ml_sequential_comparison.tex
\begin{table}[t]
\centering
\caption{Classical ML vs Sequential Models: Performance Gains Using CSD Features}
\label{tab:classical_vs_sequential}
\small
\begin{tabular}{llcccc}
\toprule
\textbf{Dataset} & \textbf{LLM} & \textbf{Classical} & \textbf{Sequential} & \textbf{Gain} \\
& & \textbf{F1 (Model)} & \textbf{F1 (Model)} & \textbf{(pts)} \\
\midrule
Math-500 & MathStral & 0.77 (SVM) & \textbf{0.88} (NN) & +11 \\
Math-500 & Qwen & 0.75 (SVM) & \textbf{0.81} (LSTM) & +6 \\
\midrule
AIME & MathStral & 0.57 (LR) & \textbf{0.77} (LSTM) & +20 \\
AIME & Qwen & 0.57 (SVM) & \textbf{0.86} (NN) & +29 \\
\midrule
MedQA & Qwen & 0.59 (LR) & \textbf{0.70} (LSTM) & +11 \\
MedQA & QwenMed & 0.59 (LR) & \textbf{0.63} (LSTM) & +4 \\
\midrule
\multicolumn{4}{l}{\textit{Average Improvement:}} & \textbf{+13.5} \\
\bottomrule
\end{tabular}
\end{table}

%% file: tables/ml_hyperparameter_csd_sfc.tex
\begin{table*}[t]
\centering
\caption{Best-performing machine learning classifiers and hyperparameters for CSD and SFC across datasets.}
\label{tab:classical_hyper_csd_sfc}

\footnotesize
\setlength{\tabcolsep}{9pt}
\renewcommand{\arraystretch}{1.25} % row spacing
\begin{tabular}{lll p{4cm} p{4cm}}
\toprule
\textbf{Dataset} & \textbf{LLM} & \textbf{Model} & \textbf{CSD Hyperparameters} & \textbf{SFC Hyperparameters} \\
\midrule
Math-500
& MathStral-7B
& SVM
& kernel=linear, $C{=}0.08$
& kernel=linear, $C{=}0.08$ \\

Math-500
& Qwen2.5-7B
& XGB
& lr$=0.07$, estimators$=200$
& lr$=0.07$, estimators$=200$ \\
\hline
Math-AIME
& MathStral-7B
& LR
& $C{=}10$, solver=lbfgs
& $C{=}10$, solver=lbfgs \\

Math-AIME
& Qwen2.5-7B
& SVM
& kernel=poly, $C{=}3$, $\gamma{=}0.07$
& kernel=poly, $C{=}3$, $\gamma{=}0.07$ \\
\hline
Med-QA
& Qwen-7B
& LR
& $C{=}7$
& $C{=}7$ \\

Med-QA
& Qwen-Med-7B
& LR
& $C{=}1\mathrm{e}{-4}$, solver=lbfgs
& $C{=}1\mathrm{e}{-4}$, solver=lbfgs \\
\bottomrule
\end{tabular}
\end{table*}

%% file: tables/seq_hyperparameter_csd_sfc.tex
\begin{table*}[!t]
\centering
\caption{Best-performing sequential models and tuned hyperparameters for CSD and SFC across datasets.}
\label{tab:sequential_hyper_csd_sfc}
\footnotesize
\renewcommand{\arraystretch}{1.25}
\setlength{\tabcolsep}{9pt}
\begin{tabular}{|l|l|l|ccc|ccc|}
\hline
\multirow{3}{*}{Dataset} &
\multirow{3}{*}{LLM} &
\multirow{2}{*}{Model} &
\multicolumn{3}{c|}{\textbf{CSD}} &
\multicolumn{3}{c|}{\textbf{SFC}} \\
\cline{4-9}
 &  &  & LR & $l_2$ & Dropout & LR & $l_2$ & Dropout \\
\hline
Math-500 & MathStral-7B & NN
& 5e--4 & 0.1 & 0.3
& 1e--4 & 0.001 & 0.2 \\
\hline
Math-500 & Qwen2.5-7B & LSTM
& 5e--3 & 0.001 & 0.2
& 5e--3 & 0.1 & 0.1 \\
\hline
Math-AIME & MathStral-7B & LSTM
& 5e--3 & 0.001 & 0.2
& 1e--4 & 0.1 & 0.01 \\
\hline
Math-AIME & Qwen2.5-7B & NN
& 1e--4 & 0.1 & 0.2
& 1e--3 & 0.1 & 0.3 \\
\hline
Med-QA & Qwen-7B & LSTM
& 1e--3 & 1e--5 & 0.2
& 1e--3 & 0.01 & 0.1 \\
\hline
Med-QA & Qwen-Med-7B & LSTM
& 5e--4 & 0.1 & 0.3
& 5e--4 & 0.001 & 0.3 \\
\hline
\end{tabular}
\end{table*}